\documentclass[letterpaper]{article}
\usepackage[preprint]{aaai2027}
\usepackage[hyphens]{url}
\usepackage{graphicx}
\usepackage{natbib}
\usepackage{caption}
\renewcommand{\dbltopfraction}{0.95}
\renewcommand{\textfraction}{0.05}
\usepackage[utf8]{inputenc}
\usepackage{booktabs}
\usepackage{amsfonts}
\usepackage{amsmath,amssymb}
\usepackage{multirow}
\usepackage{enumitem}
\usepackage{xcolor}
\usepackage{pgfplots}
\usepgfplotslibrary{groupplots}
\pgfplotsset{compat=1.18}

\definecolor{bestcolor}{RGB}{196,48,43}
\definecolor{secondcolor}{RGB}{54,102,178}
\definecolor{thirdcolor}{RGB}{52,145,78}
\definecolor{insetred}{RGB}{205,45,45}
\newcommand{\best}[1]{\textbf{\textcolor{bestcolor}{#1}}}
\newcommand{\second}[1]{\textbf{\textcolor{secondcolor}{#1}}}
\newcommand{\third}[1]{\textbf{\textcolor{thirdcolor}{#1}}}
\newcommand{\fingersize}{512}
\newcommand{\fingerroot}{fingers/iconface_qualitative/finger_\fingersize}
\newcommand{\qslot}[2]{\begin{minipage}[t]{#1}\vspace{0pt}\centering #2\end{minipage}}
\newcommand{\qhead}[2]{\qslot{#1}{{\scriptsize #2}}}
\newcommand{\qheadspaced}[2]{\qslot{#1}{{\scriptsize #2}\par\vspace{1.5pt}}}
\newcommand{\qimg}[2]{\qslot{#1}{\includegraphics[width=\linewidth]{\fingerroot/#2}}}
\newcommand{\qimgpath}[2]{%
\begin{minipage}[t]{#1}\centering
\vbox{\hbox to \linewidth{\hfil\includegraphics[width=\linewidth]{#2}\hfil}}
\end{minipage}}
\newcommand{\qimgpathinset}[3]{%
\begin{minipage}[t]{#1}\centering
\begin{tikzpicture}[baseline=(full.north)]
  \node[anchor=north,inner sep=0pt,outer sep=0pt] (full)
    {\includegraphics[width=\linewidth]{#2}};
  \node[anchor=south east,inner sep=0.45pt,outer sep=0pt,
        fill=white,draw=white,line width=0.6pt]
    at ([xshift=-1pt,yshift=1pt]full.south east) {%
      \begingroup
      \setlength{\fboxsep}{0pt}%
      \setlength{\fboxrule}{0.25pt}%
      \fcolorbox{insetred}{white}{%
        \includegraphics[width=0.40\linewidth]{#3}}%
      \endgroup};
\end{tikzpicture}
\end{minipage}}
\newcommand{\qimgarcscorepath}[3]{%
\begin{minipage}[t]{#1}\centering
\vbox{
  \hbox to \linewidth{\hfil\includegraphics[width=\linewidth]{#2}\hfil}
  \kern-10.0pt
  \hbox to \linewidth{\hfil{\begingroup\setlength{\fboxsep}{0.2pt}\colorbox{white}{\fontsize{5.6}{5.6}\selectfont Arc~#3}\endgroup}\kern0.4pt}
}
\end{minipage}}
\newcommand{\qsetlabel}[2]{\begin{minipage}[t]{#1}\vspace{0pt}\centering\vbox to \blindrowh{\vfil{\scriptsize #2}\vfil}\end{minipage}}
\newcommand{\qcaseid}[1]{\multicolumn{8}{c}{\raisebox{-1.0pt}{{\scriptsize #1}}}\\[-0.2pt]}
\newcommand{\qrefcaseid}[1]{\multicolumn{8}{c}{\raisebox{-4.5pt}{{\scriptsize #1}}}\\[-3.7pt]}
\newcommand{\datasetheader}[1]{{\small\textbf{#1}}\par\vspace{2pt}}
\newcommand{\qfigsetup}{\setlength{\tabcolsep}{1.2pt}\renewcommand{\arraystretch}{0.92}}
\newlength{\qrowsep}
\newlength{\qrowsepcompact}
\newlength{\qrefrowsepcompact}
\newlength{\qblindrowsepcompact}
\newcommand{\mainwidetabstyle}{\scriptsize\setlength{\tabcolsep}{1.0pt}\renewcommand{\arraystretch}{0.78}}
\newcommand{\methodvenue}[2]{\mbox{#1 {\fontsize{5.7}{5.7}\selectfont (#2)}}}
\newcommand{\mDMDNet}{\methodvenue{DMDNet}{TPAMI'22}}
\newcommand{\mReFLDM}{\methodvenue{ReF-LDM}{NeurIPS'24}}
\newcommand{\mInstantRestore}{\methodvenue{InstantRestore}{SIGGRAPH'25}}
\newcommand{\mFaceMe}{\methodvenue{FaceMe}{AAAI'25}}
\newcommand{\mRefSTAR}{\methodvenue{RefSTAR}{AAAI'26}}
\newcommand{\mCodeFormer}{\methodvenue{CodeFormer}{NeurIPS'22}}
\newcommand{\mGFPGAN}{\methodvenue{GFP-GAN}{CVPR'21}}
\newcommand{\mVQFR}{\methodvenue{VQFR}{ECCV'22}}
\newcommand{\mRFpp}{\methodvenue{RF++}{TPAMI'23}}
\newcommand{\mDAEFR}{\methodvenue{DAEFR}{ICLR'24}}
\newcommand{\mIConFaceOurs}{\mbox{IConFace (ours)}}
\newcommand{\teaserimg}[3]{\parbox[c][#2][c]{#1}{\centering\includegraphics[width=#1,height=#2]{#3}}}
\newcommand{\mainrefimgw}{0.110\textwidth}
\newcommand{\ablimgw}{0.06875\textwidth}
\newcommand{\blindlabelw}{0.070\textwidth}
\newcommand{\blindmainimgw}{0.123\textwidth}
\newlength{\blindrowh}
\title{IConFace: Fine-Grained Identity Conditioning for\\
Reference-Aware Face Restoration}
\author{Axi Niu\textsuperscript{*}, Jinyang Zhang\textsuperscript{*}, Senyan Qing}
\affiliations{
Northwestern Polytechnical University\\
nax@nwpu.edu.cn \quad zhangjinyang@mail.nwpu.edu.cn \quad qingsenyan@nwpu.edu.cn\\
Homepage: \url{https://cosmicrealm.github.io/IConFace/}
}

\begin{document}
\maketitle
\begingroup
\renewcommand{\thefootnote}{*}
\footnotetext{These authors contributed equally.}
\endgroup

\begin{abstract}
Severe face degradation can remove person-specific evidence, making restoration underdetermined. A generative prior may recover a sharp, plausible face yet miss localized traits that persist across images of the same person. Same-identity references supply this missing evidence, while the degraded observation anchors target structure. We propose \textbf{IConFace}, a fine-grained identity-conditioned framework that optionally conditions restoration on up to three same-identity references. Its hybrid concat backbone retains degraded and reference observations as dense visual tokens, preserving localized reference evidence. An identity pathway provides compact multi-reference guidance, while a degraded-structure pathway injects full-field and local-residual memories to reinforce target-aligned structure. We also introduce a human-audited benchmark that measures whether persistent localized identity details survive restoration. IConFace achieves leading reference compatibility, especially under severe degradation, and the highest observed preservation rate on this benchmark. Without references, it achieves leading learned perceptual quality across five blind-restoration benchmarks. Joint reference-based and paired-target evaluations show that reference-supported identity recovery and exact target agreement are complementary.
\end{abstract}

\begin{figure}[t]
    \centering
    \setlength{\tabcolsep}{1.1pt}
    \renewcommand{\arraystretch}{0.95}
    \begin{tabular}{cccc}
        {\scriptsize Ref$_1$} & {\scriptsize Ref$_2$} & {\scriptsize Ref$_3$} & {\scriptsize LQ} \\
        \teaserimg{0.225\columnwidth}{0.225\columnwidth}{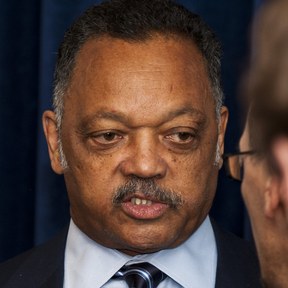} &
        \teaserimg{0.225\columnwidth}{0.225\columnwidth}{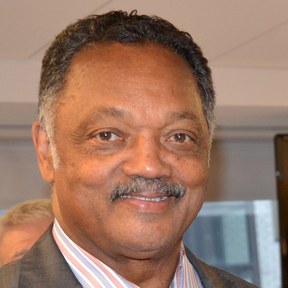} &
        \teaserimg{0.225\columnwidth}{0.225\columnwidth}{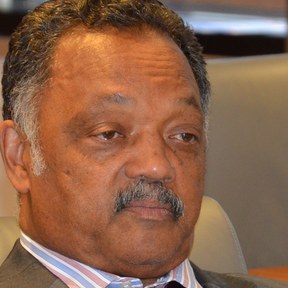} &
        \teaserimg{0.225\columnwidth}{0.225\columnwidth}{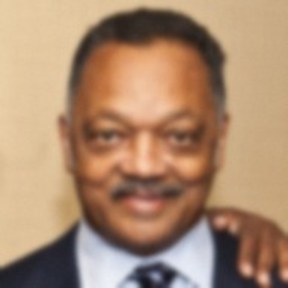} \\
        {\scriptsize CodeFormer (blind)} & {\scriptsize RefSTAR (ref.)} & {\scriptsize Ours} & {\scriptsize GT} \\
        \teaserimg{0.225\columnwidth}{0.225\columnwidth}{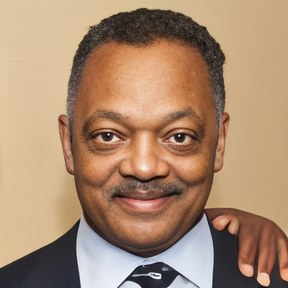} &
        \teaserimg{0.225\columnwidth}{0.225\columnwidth}{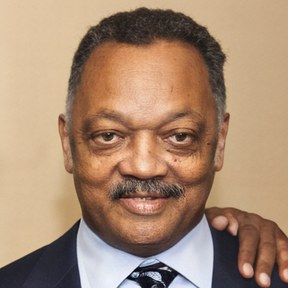} &
        \teaserimg{0.225\columnwidth}{0.225\columnwidth}{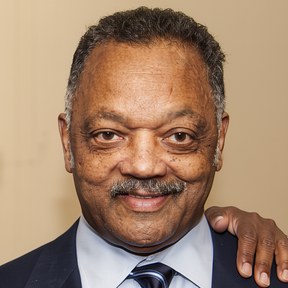} &
        \teaserimg{0.225\columnwidth}{0.225\columnwidth}{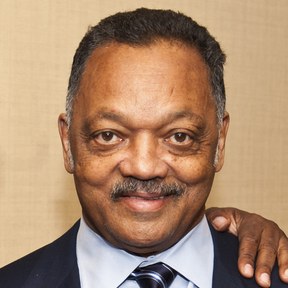} \\
        \teaserimg{0.225\columnwidth}{0.225\columnwidth}{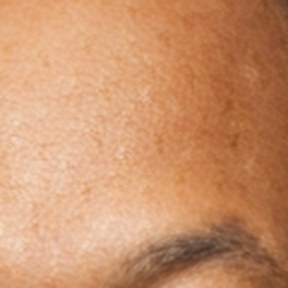} &
        \teaserimg{0.225\columnwidth}{0.225\columnwidth}{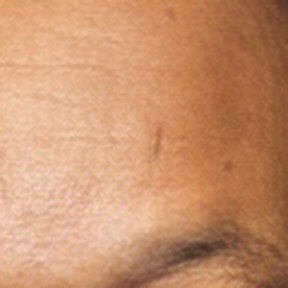} &
        \teaserimg{0.225\columnwidth}{0.225\columnwidth}{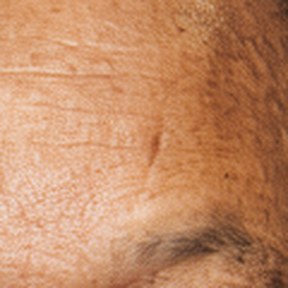} &
        \teaserimg{0.225\columnwidth}{0.225\columnwidth}{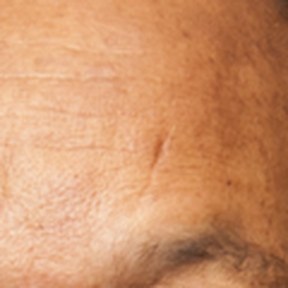} \\
        \teaserimg{0.225\columnwidth}{0.225\columnwidth}{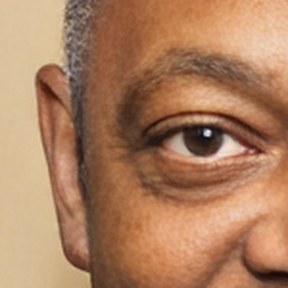} &
        \teaserimg{0.225\columnwidth}{0.225\columnwidth}{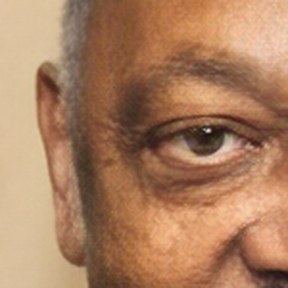} &
        \teaserimg{0.225\columnwidth}{0.225\columnwidth}{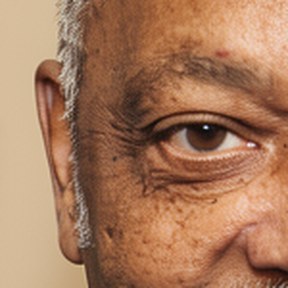} &
        \teaserimg{0.225\columnwidth}{0.225\columnwidth}{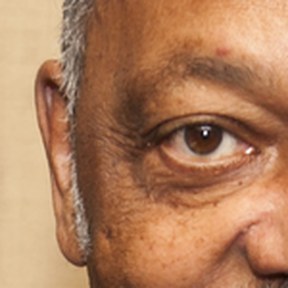}
    \end{tabular}
    \caption{\textbf{Reference-supported detail with target-aligned structure (FFHQ-Ref Moderate, case 19802).} Ref$_1$ and the paired GT show a forehead mark largely obscured in the degraded observation. IConFace recovers this cue more clearly than CodeFormer and RefSTAR while preserving GT-consistent periocular structure.}
    \label{fig:teaser}
\end{figure}

\begin{figure*}[t!]
    \centering
    \includegraphics[width=0.95\textwidth]{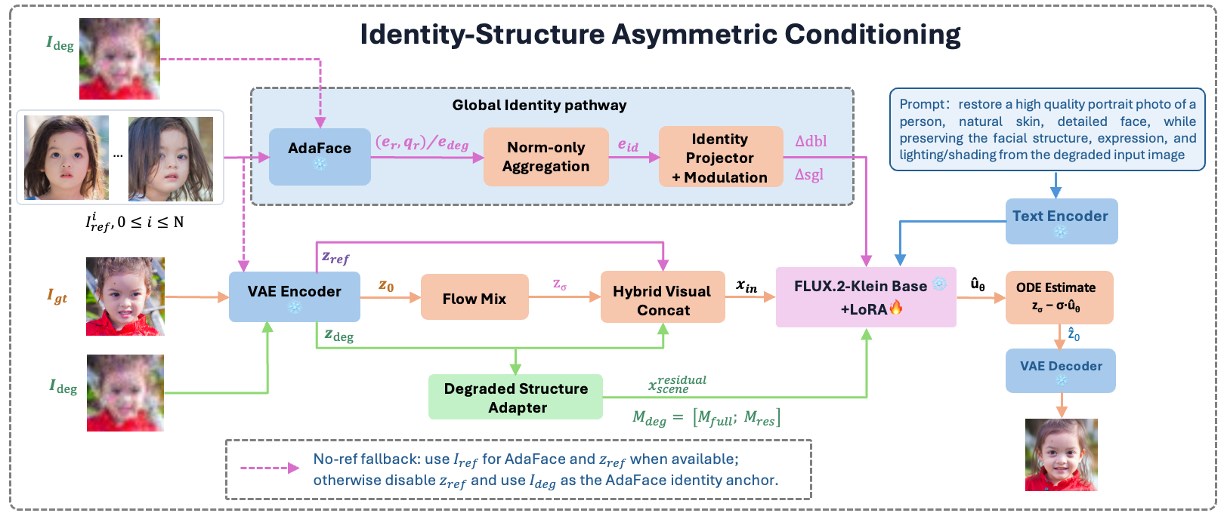}
    \caption{\textbf{IConFace framework.} The current noisy scene latent is concatenated with VAE-encoded degraded-image and optional reference tokens, keeping target-aligned degraded evidence and spatially distributed reference evidence accessible to the FLUX backbone. Norm-weighted AdaFace aggregation provides compact multi-reference identity guidance, while the degraded-structure adapter adds an input residual and full-field/local-residual memories. Purple dashed arrows denote optional routes: when references are available, they enter both the VAE and AdaFace; otherwise, the reference-token branch is omitted and the degraded image serves as the AdaFace identity anchor. FLUX predicts a clean latent for VAE decoding. Snowflakes denote frozen modules.}
    \label{fig:iconface_framework}
\end{figure*}

\section{Introduction}

Blind face restoration (BFR) recovers a high-quality face from an observation corrupted by unknown and potentially severe degradations. Low-level restoration benefits from feature and attention modeling~\cite{niu2025mpfnet,niu2024gran} and diffusion-based priors~\cite{niu2024acdmsr,tu2026tpgdiff}. For faces, generative, codebook, transformer, and diffusion priors have substantially improved sharpness and realism~\cite{menon2020pulse,wang2021gfpgan,gu2022vqfr,zhou2022codeformer,wang2022restoreformer,lin2024diffbir,miao2025flipnet}. Yet severe degradation can erase person-specific evidence: a population prior may synthesize a plausible face but cannot determine which traits were present before degradation. We focus on \emph{persistent localized identity details}, such as moles, scars, freckle patterns, and pigmentation regions that recur across captures (Fig.~\ref{fig:teaser}).

Same-identity references can supply evidence that is weak or missing in the degraded observation~\cite{li2018gfrnet,li2020asffnet,li2022dmdnet,hsiao2024refldm,liu2025faceme,chong2025refine,yin2026refstar}. Multiple captures help separate recurring identity cues from incidental appearance, but also encode capture-specific pose, expression, illumination, and texture, risking reference-state transfer. The degraded observation remains target-aligned and anchors pose, expression, and recoverable structure.

We propose \textbf{IConFace} (Fig.~\ref{fig:iconface_framework}) to combine these two evidence sources. Its hybrid concat backbone retains the degraded observation and optional references as dense visual-token segments alongside the noisy scene latent. A norm-weighted identity pathway provides compact multi-reference guidance, while a degraded-structure pathway repeatedly reinforces target-aligned evidence through an input residual and full-field/local-residual memories. The model accepts zero to three references.

Beyond perceptual and global-identity evaluation~\cite{wang2026ntireface}, reference-aware restoration must distinguish compatibility with cross-capture identity evidence from agreement with one paired target. We therefore combine RefAvg, paired-GT fidelity, and supplementary pose, expression, and parsing diagnostics to assess identity recovery and target-state retention. Since these global measures do not identify which trait survives, we also introduce a human-audited detail-preservation benchmark whose retained traits are supported by both the paired target and a same-identity reference.

Across three reference-aware benchmarks, IConFace achieves the strongest reference compatibility, with the clearest gains under severe degradation, and the highest observed localized-detail preservation rate. Without references, it achieves leading learned perceptual quality across five blind-restoration benchmarks. Our contributions are:
\begin{itemize}[leftmargin=1.15em,itemsep=1pt,topsep=2pt]
\item We propose a fine-grained multi-reference identity-conditioning design that retains references as spatially distributed visual tokens and complements them with norm-weighted global identity aggregation, preserving localized evidence.
\item We introduce a degraded-structure reinforcement pathway that injects an input residual and full-field/local-residual memories, keeping restoration anchored to the target observation.
\item We develop a human-audited localized-detail benchmark and evaluate IConFace against diverse methods across multiple datasets, showing the highest observed preservation rate, leading reference compatibility and learned perceptual quality, and clear qualitative gains in fine-grained identity recovery.
\end{itemize}

\section{Related Work}

\textbf{Blind face restoration.}
Blind face restoration reconstructs a portrait without subject-specific auxiliary images. Early approaches use facial components or learned dictionaries~\cite{li2020dfdnet}. Generative-prior methods embed pretrained face distributions into restoration networks~\cite{yang2021gpen,wang2021gfpgan}, while codebook and transformer methods retrieve or compose high-quality representations from discrete or key--value priors~\cite{gu2022vqfr,zhou2022codeformer,wang2022restoreformer,wang2023restoreformerpp}. Diffusion-based approaches further improve realism and degradation robustness~\cite{zhao2023idm,qiu2023diffbfr,yang2023pgdiff,lin2024diffbir,miao2025flipnet,wang2025osdface}. These methods condition only on the degraded observation; when corruption removes a persistent identity detail, a population prior cannot determine which subject-specific evidence is missing.

\textbf{Reference-based face restoration.}
Reference-based methods supplement the degraded observation with same-identity images. GFRNet predicts a spatial warp from one guidance image, ASFFNet selects and fuses aligned exemplars, and DMDNet stores generic and identity-specific component features~\cite{li2018gfrnet,li2020asffnet,li2022dmdnet}. Later generative approaches use per-subject personalization, key--value conditioning, identity adapters, shared attention, or compact identity prompts~\cite{varanka2024pfstorer,hsiao2024refldm,zhang2024instantrestore,liu2025faceme}; ReFine and RefSTAR emphasize detail transfer or reference feature selection~\cite{chong2025refine,yin2026refstar}. These designs range from compact global summaries to dense or selectively aligned local features. Compact summaries may omit spatially localized evidence, whereas richer transfer can import the reference capture's pose, expression, illumination, or incidental appearance. IConFace retains both input sources as dense tokens and adds asymmetric conditioning to reinforce their complementary roles.

\textbf{Identity representation and evaluation.}
Margin-based recognizers provide compact identity representations with strong global discriminability~\cite{deng2019arcface,meng2021magface,kim2022adaface}. Restoration methods reuse these embeddings for conditioning, supervision, and cosine-similarity evaluation~\cite{hsiao2024refldm,liu2025faceme,chong2025refine}. Global similarity is useful but cannot identify which localized trait was recovered or distinguish identity recovery from reference-state transfer. Paired-target metrics measure agreement with one capture, perceptual metrics measure visual quality, and target/reference state comparisons expose possible shifts toward a reference. Existing fine-detail studies still emphasize global similarity, qualitative comparison, and user preference~\cite{chong2025refine}.

\section{Method}

\subsection{Overview and Dense Evidence Backbone}
Given a degraded image $I_{\mathrm{deg}}$ and same-identity references $\mathcal{R}=\{I_{\mathrm{ref}}^1,\ldots,I_{\mathrm{ref}}^N\}$, IConFace defines
\begin{equation}
    \hat{I}=G_{\theta}(I_{\mathrm{deg}},\mathcal{R}),
    \qquad N\in\{0,1,2,3\},
\end{equation}
which operates in reference-aware mode for $N>0$ and no-reference mode for $N=0$. The degraded image is directly associated with the target observation and anchors its layout and recoverable structure; $\mathcal{R}$ supplies additional evidence for persistent identity traits.

The backbone is FLUX.2-Klein-base-4B~\cite{blackforestlabs2025flux2}. Its VAE encoder $E$, decoder $D$, and text encoder are frozen, while the transformer is adapted through LoRA~\cite{hu2022lora} together with the proposed conditioning modules. A fixed restoration prompt provides text context and is omitted below. At flow state $z_\sigma$, the scene, degraded, and reference latent grids are independently packed without interpolation or spatial resampling:
\begin{equation}
\begin{aligned}
 z_{\mathrm{deg}}&=E(I_{\mathrm{deg}}), & z_{\mathrm{ref}}^r&=E(I_{\mathrm{ref}}^r),\\
 x_{\mathrm{scene}}&=\mathrm{Pack}(z_\sigma), & x_{\mathrm{deg}}&=\mathrm{Pack}(z_{\mathrm{deg}}),\\
 x_{\mathrm{ref}}^r&=\mathrm{Pack}(z_{\mathrm{ref}}^r), &
 x_{\mathrm{in}}&=[x_{\mathrm{scene}};x_{\mathrm{deg}};x_{\mathrm{ref}}^1;\ldots;x_{\mathrm{ref}}^N].
\end{aligned}
\end{equation}
The grids have the same spatial resolution, and $\mathrm{Pack}$ only rearranges each $(B,C,H,W)$ tensor into a token sequence. Visual tokens use source-aware 3D RoPE to preserve spatial layout and distinguish the scene, degraded observation, and reference groups; details are provided in the supplement. References remain dense spatial token sequences rather than being represented solely by a global identity vector; for $N=0$, the reference segment is omitted. Dense concatenation exposes both evidence sources to the backbone but does not prescribe their roles, motivating the two asymmetric pathways below.

\begin{figure*}[t]
    \centering
    \datasetheader{Reference-aware qualitative comparisons}
    \begingroup
    \qfigsetup
    \begin{tabular}{@{}cccccccc@{}}
    \qheadspaced{\mainrefimgw}{Reference} &
    \qheadspaced{\mainrefimgw}{LQ} &
    \qheadspaced{\mainrefimgw}{ReF-LDM} &
    \qheadspaced{\mainrefimgw}{InstantR.} &
    \qheadspaced{\mainrefimgw}{FaceMe} &
    \qheadspaced{\mainrefimgw}{RefSTAR} &
    \qheadspaced{\mainrefimgw}{Ours} &
    \qheadspaced{\mainrefimgw}{GT}\\[-1pt]
    \qrefcaseid{CelebA-Test-Ref: case 00645}
    \qimgpathinset{\mainrefimgw}{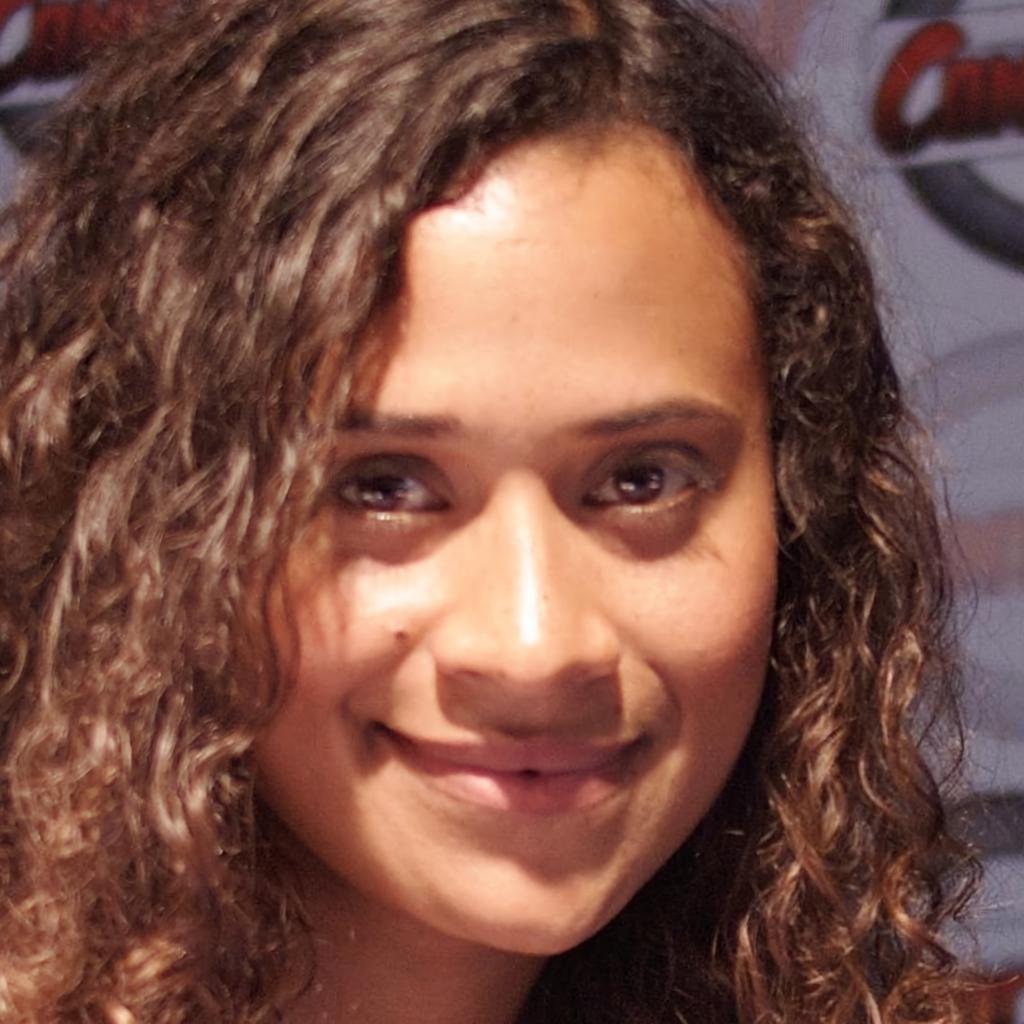}{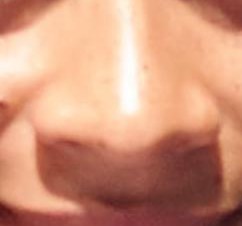} &
    \qimgpathinset{\mainrefimgw}{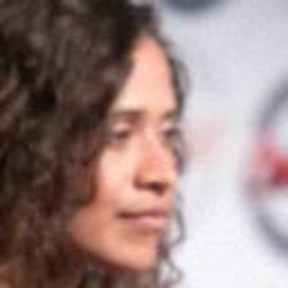}{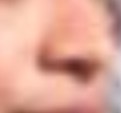} &
    \qimgpathinset{\mainrefimgw}{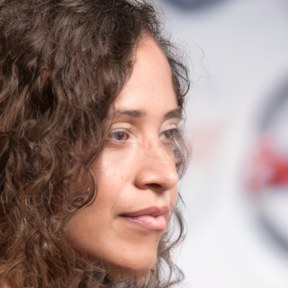}{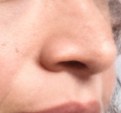} &
    \qimgpathinset{\mainrefimgw}{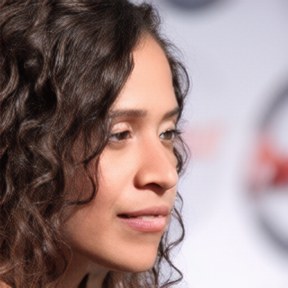}{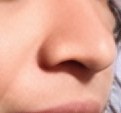} &
    \qimgpathinset{\mainrefimgw}{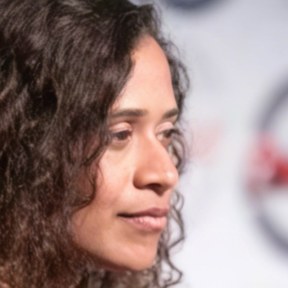}{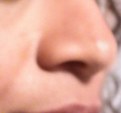} &
    \qimgpathinset{\mainrefimgw}{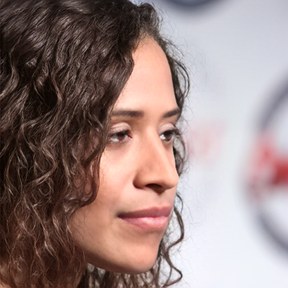}{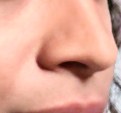} &
    \qimgpathinset{\mainrefimgw}{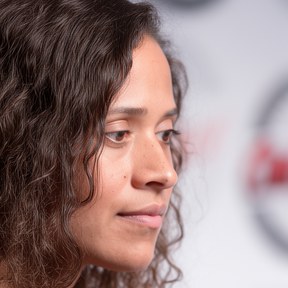}{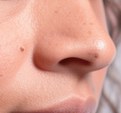} &
    \qimgpathinset{\mainrefimgw}{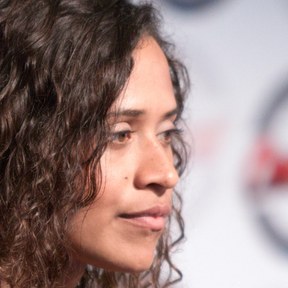}{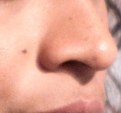}\\[\qrefrowsepcompact]
    \qrefcaseid{FFHQ-Ref Moderate: case 03016}
    \qimgpathinset{\mainrefimgw}{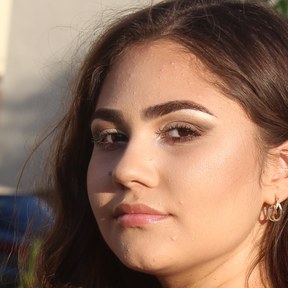}{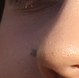} &
    \qimgpathinset{\mainrefimgw}{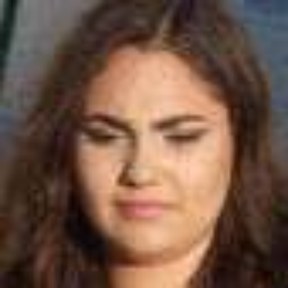}{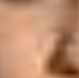} &
    \qimgpathinset{\mainrefimgw}{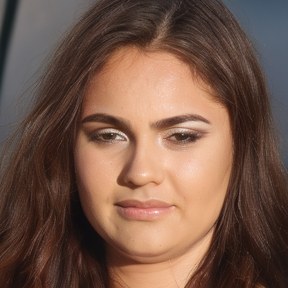}{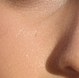} &
    \qimgpathinset{\mainrefimgw}{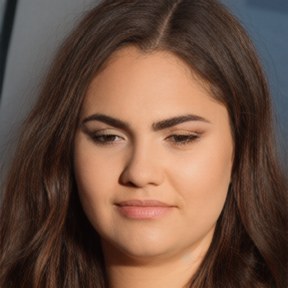}{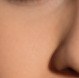} &
    \qimgpathinset{\mainrefimgw}{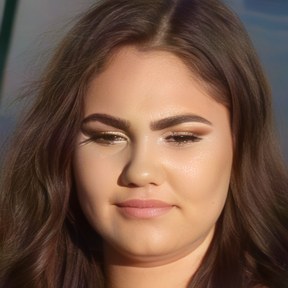}{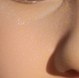} &
    \qimgpathinset{\mainrefimgw}{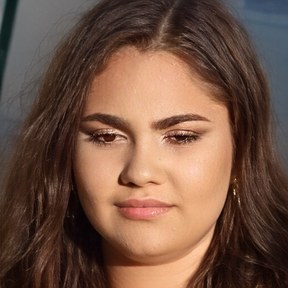}{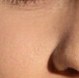} &
    \qimgpathinset{\mainrefimgw}{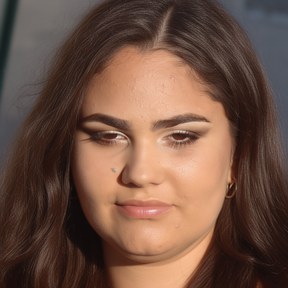}{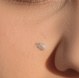} &
    \qimgpathinset{\mainrefimgw}{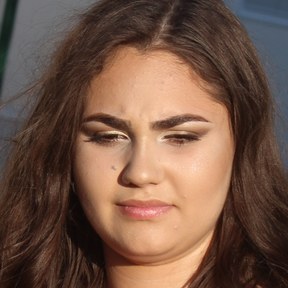}{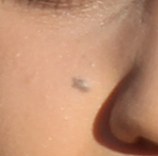}\\[\qrefrowsepcompact]
    \qrefcaseid{FFHQ-Ref Severe: case 12744}
    \qimgpathinset{\mainrefimgw}{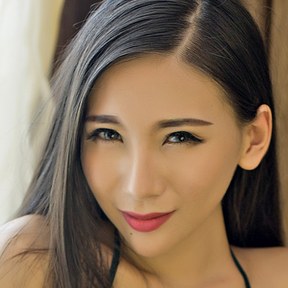}{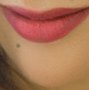} &
    \qimgpathinset{\mainrefimgw}{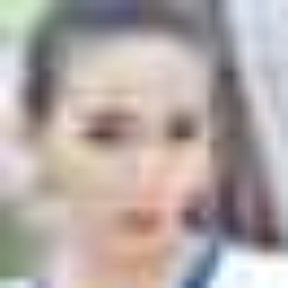}{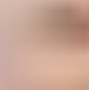} &
    \qimgpathinset{\mainrefimgw}{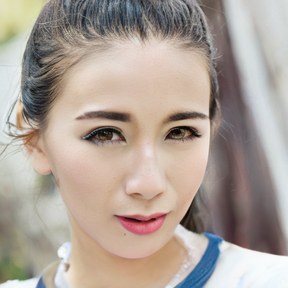}{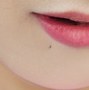} &
    \qimgpathinset{\mainrefimgw}{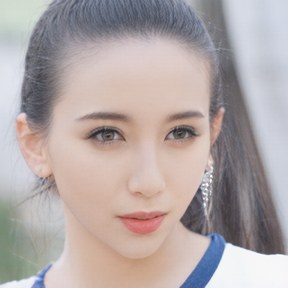}{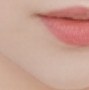} &
    \qimgpathinset{\mainrefimgw}{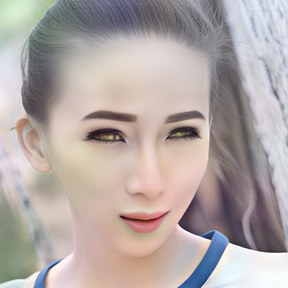}{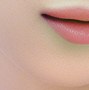} &
    \qimgpathinset{\mainrefimgw}{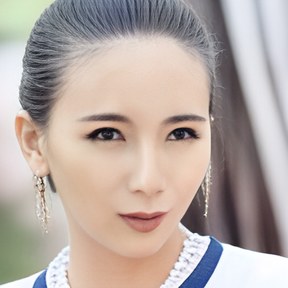}{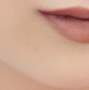} &
    \qimgpathinset{\mainrefimgw}{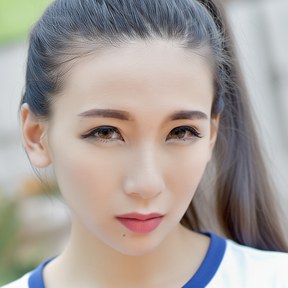}{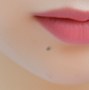} &
    \qimgpathinset{\mainrefimgw}{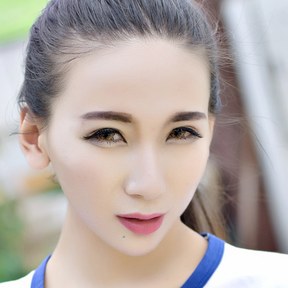}{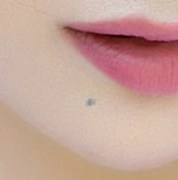}\\[\qrefrowsepcompact]
    \end{tabular}
    \endgroup
    \caption{\textbf{Reference-aware qualitative comparisons on three benchmarks.} Each row shows one displayed reference, LQ, restoration outputs, and the paired GT; bottom-right insets enlarge a row-matched facial region. Methods use their official reference configurations, with one reference displayed when more are used.}
    \label{fig:qual_ref_compact}
\end{figure*}

\subsection{Asymmetric Identity-Structure Conditioning}
\paragraph{Multi-reference identity guidance.}
For each available reference $r\in\mathcal{V}$, frozen AdaFace $A$~\cite{kim2022adaface} produces $a_r\in\mathbb{R}^{512}$, decomposed as
\begin{equation}
    e_r=\frac{a_r}{\lVert a_r\rVert_2},\qquad
    q_r=\lVert a_r\rVert_2.
\end{equation}
Following hyperspherical face recognition~\cite{deng2019arcface,kim2022adaface,meng2021magface}, $e_r$ represents angular identity evidence and the pre-normalization magnitude $q_r$ serves as a relative recognizability proxy, not a calibrated probability. We use these magnitudes to weight the same-identity directions and select the effective identity anchor:
\begin{equation}
\begin{aligned}
    w_r &= \frac{\exp(\log q_r/T)}
    {\sum_{j\in\mathcal{V}}\exp(\log q_j/T)},\\
    e_{\mathrm{ref}} &= \mathrm{Norm}\!\left(
    \sum_{r\in\mathcal{V}}w_r e_r\right),\\
 e_{\mathrm{id}}&=
 \begin{cases}
 e_{\mathrm{ref}}, & \mathcal{V}\neq\emptyset,\\
 \mathrm{Norm}(A(I_{\mathrm{deg}})), & \mathcal{V}=\emptyset.
 \end{cases}
\end{aligned}
\end{equation}
The degraded fallback preserves the conditioning interface at $N=0$ and is used only for forward conditioning, never as a reference supervision target. For transformer stream $s\in\{\mathrm{dbl},\mathrm{sgl}\}$, the projected anchor predicts image-modulation deltas:
\begin{equation}
\begin{aligned}
    h_{\mathrm{id}}&=\phi_{\mathrm{id}}(e_{\mathrm{id}}),\\
    \Delta^{s}&=\psi^{s}(h_{\mathrm{id}})
    =(\Delta^{s,\mathrm{shift}},\Delta^{s,\mathrm{scale}},\Delta^{s,\mathrm{gate}}),\\
    m_\ell'&=m_\ell+\boldsymbol{\gamma}_\ell\odot\Delta^{s(\ell)}.
\end{aligned}
\end{equation}
The learned positive scales $\boldsymbol{\gamma}_\ell$ control the three modulation components at each image block. This pathway does not modify the text stream. The compact embedding supplies global identity guidance, while dense reference tokens remain available for localized evidence that the embedding may not retain.

\paragraph{Degraded-structure anchoring.}
The degraded image is the only input aligned with the target observation. Because $x_{\mathrm{deg}}$ and $x_{\mathrm{scene}}$ share a spatial grid, we inject a low-rank input residual into the scene stream and form a two-route compressed memory:
\begin{equation}
\begin{aligned}
    x_{\mathrm{scene}}'&=x_{\mathrm{scene}}+s_{\mathrm{deg}}W_{\mathrm{in}}(x_{\mathrm{deg}}),\\
    H_{\mathrm{deg}} &= W_m(x_{\mathrm{deg}}),\\
    M_{\mathrm{full}} &= P_f(H_{\mathrm{deg}}),\\
    M_{\mathrm{res}} &= P_r(H_{\mathrm{deg}}-
        \mathrm{Smooth}(H_{\mathrm{deg}})),\\
    M_{\mathrm{deg}} &= [M_{\mathrm{full}};M_{\mathrm{res}}].
\end{aligned}
\end{equation}
Here $s_{\mathrm{deg}}$ is a fixed structure strength, and $P_f,P_r$ are learned-query poolers. The full-field route summarizes the complete projected field, while the local-residual route emphasizes localized deviations after fixed smoothing. Together, they retain complementary spatial scales of target-aligned structure.

Each image block reads the memory through stream-specific low-rank K/V projections:
\begin{equation}
    h_\ell'=h_\ell+s_{\mathrm{deg}}
    \mathrm{Attn}(Q_\ell h_\ell,
    K_\ell M_{\mathrm{deg}},V_\ell M_{\mathrm{deg}}).
\end{equation}
Together, the input residual and block-wise memory keep target-aligned degraded evidence accessible throughout denoising, while the dense reference tokens and global identity pathway provide complementary identity information.

\subsection{Training Objectives and Inference}
For clean target latent $z_0=E(I_{\mathrm{gt}})$, noise $\epsilon\sim\mathcal{N}(0,I)$, and interpolation level $\sigma\in[0,1]$, flow matching~\cite{lipman2023flowmatching} uses
\begin{equation}
\begin{aligned}
    z_\sigma&=(1-\sigma)z_0+\sigma\epsilon,\\
    u^\star&=\epsilon-z_0,\\
    \hat{u}_\theta&=F_\theta(z_\sigma,\sigma;
    I_{\mathrm{deg}},\mathcal{R}),\\
    \mathcal{L}_{\mathrm{fm}}&=
    \mathbb{E}_{z_0,\epsilon,\sigma}
    \left[\lVert\hat{u}_\theta-(\epsilon-z_0)\rVert_2^2\right].
\end{aligned}
\end{equation}

The flow-matching objective regresses the target velocity uniformly over timesteps. For reference-aware samples, frozen AdaFace additionally constrains the decoded clean-latent estimate:
\begin{equation}
\begin{aligned}
    \tilde{z}_0&=z_\sigma-\sigma\hat{u}_\theta,\qquad \tilde{I}=D(\tilde{z}_0),\\
    \mathcal{L}_{\mathrm{ref\mbox{-}id}}
    &=1-\cos(A(\tilde{I}),\mathrm{sg}(e_{\mathrm{ref}})),\\
    \mathcal{L}_{\mathrm{gt\mbox{-}id}}
    &=1-\cos(A(\tilde{I}),\mathrm{sg}(e_{\mathrm{gt}})),
\end{aligned}
\end{equation}
where $e_{\mathrm{gt}}=\mathrm{Norm}(A(I_{\mathrm{gt}}))$ and $\mathrm{sg}$ stops gradients through the identity targets. An adaptive clean-target stabilizer increases with reference--target embedding disagreement:
\begin{equation}
\begin{aligned}
    \lambda_h^\star &= \lambda_h\,
    \mathrm{clamp}(1-\cos(e_{\mathrm{ref}},e_{\mathrm{gt}}),0,1),\\
    \ell_{\mathrm{id}} &= \omega(\sigma)\left[
    (1-\lambda_h^\star)\mathcal{L}_{\mathrm{ref\mbox{-}id}}
    +\lambda_h^\star\mathcal{L}_{\mathrm{gt\mbox{-}id}}
    \right],
\end{aligned}
\end{equation}
where $\lambda_h=0.25$, $\omega(\sigma)=\max(1-\sigma,0.25)^2$, and $\ell_{\mathrm{id}}=0$ when $\mathcal{V}=\emptyset$. The timestep weight reduces identity supervision at high-noise states rather than hard-masking them. With $\mathcal{L}_{\mathrm{id}}=\mathbb{E}[\ell_{\mathrm{id}}]$,
\begin{equation}
    \mathcal{L}=\alpha_{\mathrm{fm}}\mathcal{L}_{\mathrm{fm}}
    +\lambda_{\mathrm{id}}\mathcal{L}_{\mathrm{id}},
\end{equation}
with $\alpha_{\mathrm{fm}}=0.75$ and $\lambda_{\mathrm{id}}=0.30$.

Training varies reference availability from zero to three. At inference, the predicted velocity field is integrated over the sampling schedule to obtain $\hat{z}_0$, which is decoded as $\hat{I}=D(\hat{z}_0)$. The supplement reports reference sampling, token budgets, low-rank dimensions, optimization settings, and a controlled reference-count diagnostic.

\begin{figure}[!b]
    \centering
    \includegraphics[width=0.60\columnwidth]{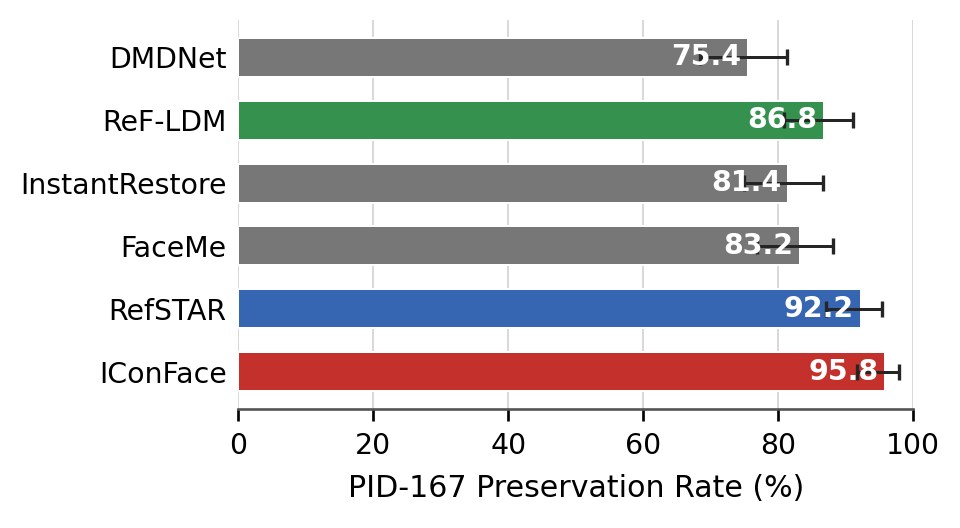}
    \caption{\textbf{Localized-detail preservation.} Rates over 167 audited traits supported by the paired GT and a same-identity reference; error bars indicate estimation uncertainty.}
    \label{fig:main_pid_preservation_rate}
\end{figure}

\begin{figure*}[t]
    \centering
    \includegraphics[width=0.80\textwidth]{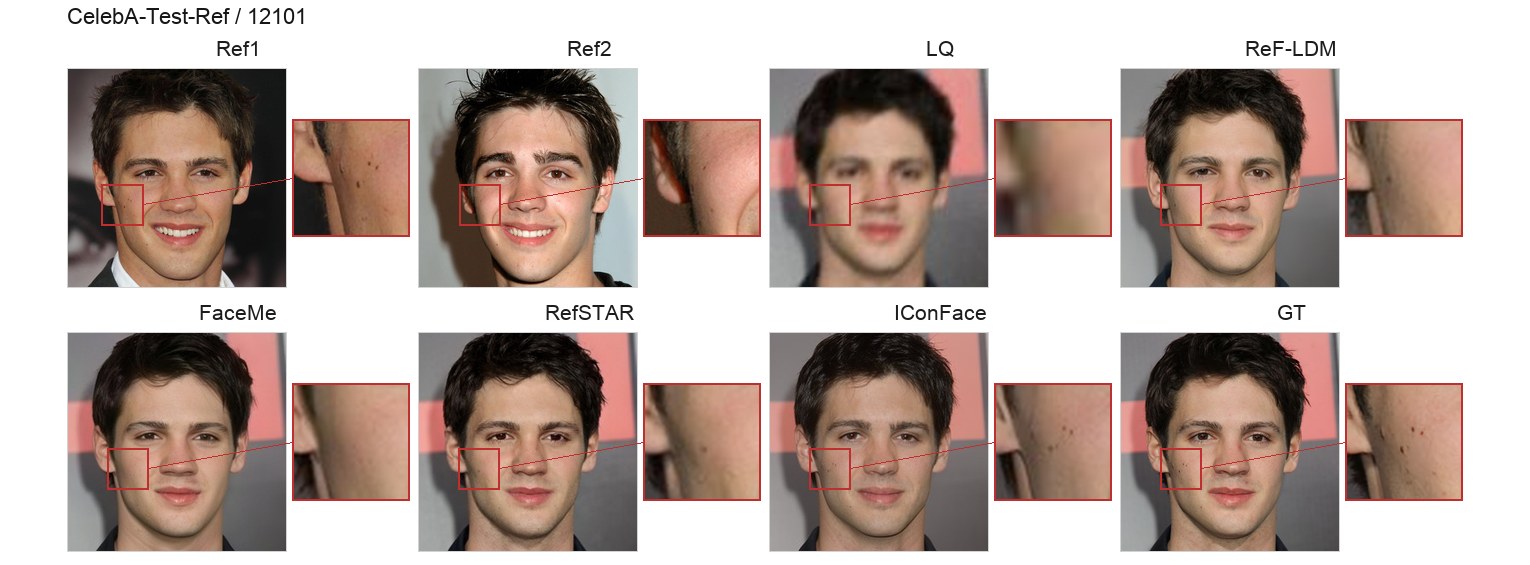}
    \caption{\textbf{Localized-detail preservation.} IConFace preserves the three-mole cheek pattern and its spatial arrangement more completely than competing restorations, which omit or weaken some moles. Insets enlarge the region.}
    \label{fig:main_visible_feature_examples}
\end{figure*}

\begin{figure*}[t]
    \centering
    \begingroup
    \setlength{\tabcolsep}{1.0pt}
    \renewcommand{\arraystretch}{0.92}
    \resizebox{0.55\textwidth}{!}{%
    \begin{tabular}{@{}cccccccc@{}}
    \qheadspaced{\blindlabelw}{Set} &
    \qheadspaced{\blindmainimgw}{LQ} &
    \qheadspaced{\blindmainimgw}{CodeFormer} &
    \qheadspaced{\blindmainimgw}{GFP-GAN} &
    \qheadspaced{\blindmainimgw}{VQFR} &
    \qheadspaced{\blindmainimgw}{RF++} &
    \qheadspaced{\blindmainimgw}{DAEFR} &
    \qheadspaced{\blindmainimgw}{IConFace}\\[-1pt]
    \qsetlabel{\blindlabelw}{LFW} &
    \qimg{\blindmainimgw}{LFW/Adrian_Annus_0001_00__deg.jpg} &
    \qimg{\blindmainimgw}{LFW/Adrian_Annus_0001_00__codeformer.jpg} &
    \qimg{\blindmainimgw}{LFW/Adrian_Annus_0001_00__gfpgan.jpg} &
    \qimg{\blindmainimgw}{LFW/Adrian_Annus_0001_00__vqfr.jpg} &
    \qimg{\blindmainimgw}{LFW/Adrian_Annus_0001_00__restoreformerpp.jpg} &
    \qimg{\blindmainimgw}{LFW/Adrian_Annus_0001_00__daefr.jpg} &
    \qimg{\blindmainimgw}{LFW/Adrian_Annus_0001_00__ours.jpg}\\[\qblindrowsepcompact]
    \qsetlabel{\blindlabelw}{CelebChild} &
    \qimg{\blindmainimgw}{CelebChild/Child__040_Zooey_Deschanel_00__deg.jpg} &
    \qimg{\blindmainimgw}{CelebChild/Child__040_Zooey_Deschanel_00__codeformer.jpg} &
    \qimg{\blindmainimgw}{CelebChild/Child__040_Zooey_Deschanel_00__gfpgan.jpg} &
    \qimg{\blindmainimgw}{CelebChild/Child__040_Zooey_Deschanel_00__vqfr.jpg} &
    \qimg{\blindmainimgw}{CelebChild/Child__040_Zooey_Deschanel_00__restoreformerpp.jpg} &
    \qimg{\blindmainimgw}{CelebChild/Child__040_Zooey_Deschanel_00__daefr.jpg} &
    \qimg{\blindmainimgw}{CelebChild/Child__040_Zooey_Deschanel_00__ours.jpg}\\[\qblindrowsepcompact]
    \qsetlabel{\blindlabelw}{WebPhoto} &
    \qimg{\blindmainimgw}{WebPhoto/00022_00__deg.jpg} &
    \qimg{\blindmainimgw}{WebPhoto/00022_00__codeformer.jpg} &
    \qimg{\blindmainimgw}{WebPhoto/00022_00__gfpgan.jpg} &
    \qimg{\blindmainimgw}{WebPhoto/00022_00__vqfr.jpg} &
    \qimg{\blindmainimgw}{WebPhoto/00022_00__restoreformerpp.jpg} &
    \qimg{\blindmainimgw}{WebPhoto/00022_00__daefr.jpg} &
    \qimg{\blindmainimgw}{WebPhoto/00022_00__ours.jpg}\\[\qblindrowsepcompact]
    \qsetlabel{\blindlabelw}{Wider-Test} &
    \qimg{\blindmainimgw}{Wider-Test/0060__deg.jpg} &
    \qimg{\blindmainimgw}{Wider-Test/0060__codeformer.jpg} &
    \qimg{\blindmainimgw}{Wider-Test/0060__gfpgan.jpg} &
    \qimg{\blindmainimgw}{Wider-Test/0060__vqfr.jpg} &
    \qimg{\blindmainimgw}{Wider-Test/0060__restoreformerpp.jpg} &
    \qimg{\blindmainimgw}{Wider-Test/0060__daefr.jpg} &
    \qimg{\blindmainimgw}{Wider-Test/0060__ours.jpg}\\[\qblindrowsepcompact]
    \qsetlabel{\blindlabelw}{CelebA-Test} &
    \qimg{\blindmainimgw}{CelebA-Test/00000216__deg.jpg} &
    \qimg{\blindmainimgw}{CelebA-Test/00000216__codeformer.jpg} &
    \qimg{\blindmainimgw}{CelebA-Test/00000216__gfpgan.jpg} &
    \qimg{\blindmainimgw}{CelebA-Test/00000216__vqfr.jpg} &
    \qimg{\blindmainimgw}{CelebA-Test/00000216__restoreformerpp.jpg} &
    \qimg{\blindmainimgw}{CelebA-Test/00000216__daefr.jpg} &
    \qimg{\blindmainimgw}{CelebA-Test/00000216__ours.jpg}\\
    \end{tabular}
    }
    \endgroup
    \caption{\textbf{No-reference comparisons on five benchmarks.} Rows show LQ and six outputs. Across varied ages, poses, occlusions, and illumination, IConFace restores sharp, coherent facial structure and local detail.}
    \label{fig:qual_noref_compact}
    \par\vspace{4pt}
    \begingroup
    \qfigsetup
    \begin{tabular}{@{}cccccccc@{}}
    \qheadspaced{\ablimgw}{Ref} &
    \qheadspaced{\ablimgw}{LQ} &
    \qheadspaced{\ablimgw}{Concat} &
    \qheadspaced{\ablimgw}{+ Struct} &
    \qheadspaced{\ablimgw}{+ ID} &
    \qheadspaced{\ablimgw}{\shortstack{ID+Struct\\(1R)}} &
    \qheadspaced{\ablimgw}{Full (2R)} &
    \qheadspaced{\ablimgw}{GT}\\[-1pt]
    \qcaseid{FFHQ-Ref Moderate: case 03234}
    \qimgarcscorepath{\ablimgw}{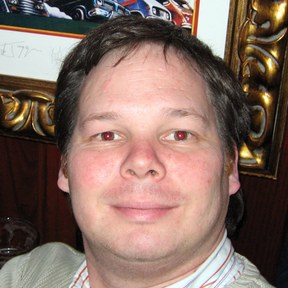}{1.000} &
    \qimgarcscorepath{\ablimgw}{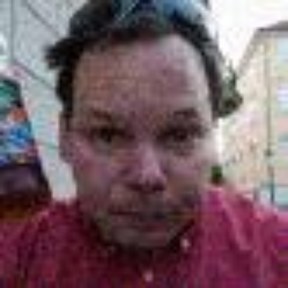}{0.479} &
    \qimgarcscorepath{\ablimgw}{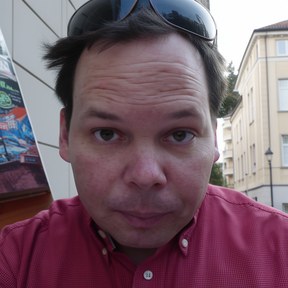}{0.553} &
    \qimgarcscorepath{\ablimgw}{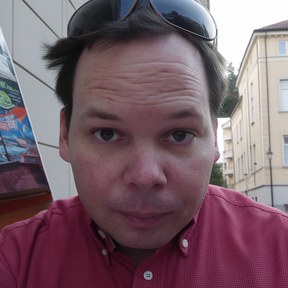}{0.602} &
    \qimgarcscorepath{\ablimgw}{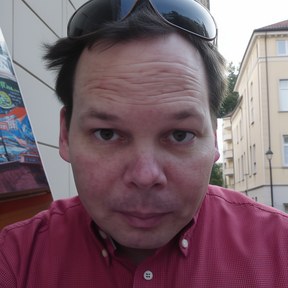}{0.643} &
    \qimgarcscorepath{\ablimgw}{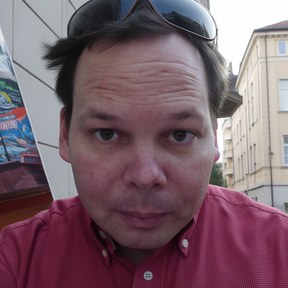}{0.648} &
    \qimgarcscorepath{\ablimgw}{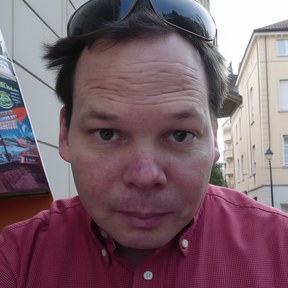}{0.656} &
    \qimgarcscorepath{\ablimgw}{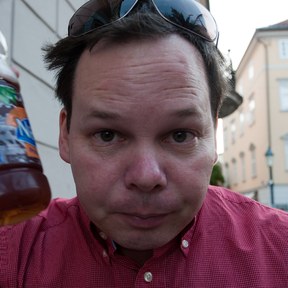}{0.607}\\[\qrowsepcompact]
    \qcaseid{FFHQ-Ref Severe: case 31223}
    \qimgarcscorepath{\ablimgw}{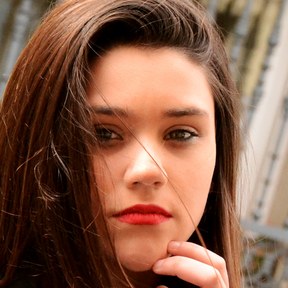}{1.000} &
    \qimgarcscorepath{\ablimgw}{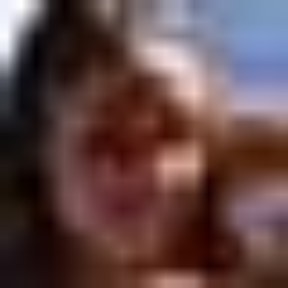}{0.047} &
    \qimgarcscorepath{\ablimgw}{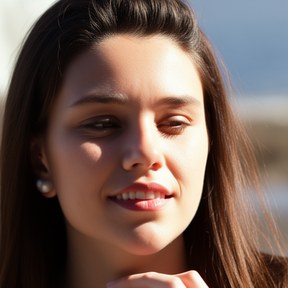}{0.467} &
    \qimgarcscorepath{\ablimgw}{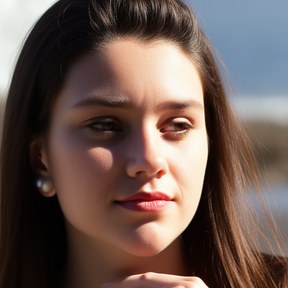}{0.523} &
    \qimgarcscorepath{\ablimgw}{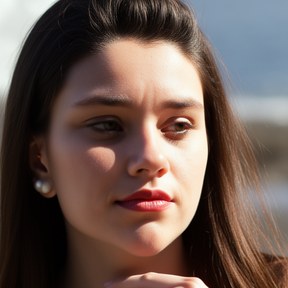}{0.535} &
    \qimgarcscorepath{\ablimgw}{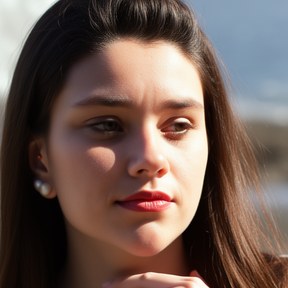}{0.540} &
    \qimgarcscorepath{\ablimgw}{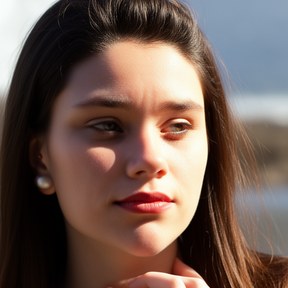}{0.588} &
    \qimgarcscorepath{\ablimgw}{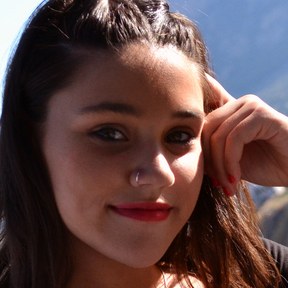}{0.545}\\
    \end{tabular}
    \endgroup
    \caption{\textbf{Qualitative component ablations.} Arc labels compare with displayed Ref$_1$; headline identity metrics use all protocol references.}
    \label{fig:iconface_ablation_ref_compact}
\end{figure*}

\begin{table*}[t]
    \centering
    \caption{Reference-aware comparison under each method's official reference configuration; RefAvg uses only its supplied references.}
    \label{tab:main_reference_results}
    {\mainwidetabstyle
    \begin{tabular}{@{}llccccc@{}}
        \toprule
        Method & Dataset & RefAvg$_{\mathrm{ArcFace}}$ $\uparrow$ & RefAvg$_{\mathrm{MagFace}}$ $\uparrow$ & MUSIQ $\uparrow$ & CLIP-IQA $\uparrow$ & MANIQA $\uparrow$ \\
        \midrule
\mDMDNet & \multirow{6}{*}{\shortstack{CelebA-Test\\Ref}} & 0.462 & 0.519 & \third{72.725} & 0.625 & 0.495 \\
\mReFLDM & & \third{0.561} & \third{0.616} & \second{74.426} & \second{0.675} & \second{0.527} \\
\mInstantRestore & & 0.522 & 0.583 & 71.324 & 0.555 & 0.494 \\
\mFaceMe & & 0.534 & 0.591 & 70.716 & 0.586 & 0.480 \\
\mRefSTAR & & \second{0.594} & \second{0.652} & 71.633 & \third{0.631} & \third{0.509} \\
        \mIConFaceOurs & & \best{0.633} & \best{0.678} & \best{76.068} & \best{0.727} & \best{0.635} \\
        \midrule
\mDMDNet & \multirow{6}{*}{\shortstack{FFHQ-Ref\\Moderate}} & 0.567 & 0.610 & 72.024 & 0.633 & 0.477 \\
\mReFLDM & & 0.634 & 0.671 & \second{75.254} & \second{0.708} & \second{0.544} \\
\mInstantRestore & & 0.600 & 0.643 & 69.358 & 0.568 & 0.463 \\
\mFaceMe & & \third{0.635} & \third{0.674} & 72.480 & 0.610 & 0.491 \\
\mRefSTAR & & \second{0.648} & \second{0.686} & \third{72.890} & \third{0.646} & \third{0.501} \\
        \mIConFaceOurs & & \best{0.688} & \best{0.717} & \best{76.220} & \best{0.738} & \best{0.613} \\
        \midrule
\mDMDNet & \multirow{6}{*}{\shortstack{FFHQ-Ref\\Severe}} & 0.137 & 0.174 & 62.177 & 0.543 & 0.355 \\
\mReFLDM & & \second{0.588} & \third{0.624} & \best{75.744} & \second{0.717} & \second{0.552} \\
\mInstantRestore & & 0.467 & 0.516 & 69.998 & 0.581 & 0.467 \\
\mFaceMe & & 0.455 & 0.507 & 72.251 & 0.610 & 0.493 \\
\mRefSTAR & & \third{0.580} & \second{0.628} & \third{73.012} & \third{0.645} & \third{0.512} \\
        \mIConFaceOurs & & \best{0.682} & \best{0.713} & \second{75.679} & \best{0.726} & \best{0.600} \\
        \bottomrule
    \end{tabular}}
\end{table*}

\begin{table*}[t]
    \centering
    \caption{Paired-GT similarity to protocol references. Similarities are averaged per sample; Moderate and Severe share the FFHQ-Ref pairs.}
    \label{tab:main_ref_gt_gap}
    {\scriptsize\setlength{\tabcolsep}{2.2pt}\renewcommand{\arraystretch}{0.78}
    \begin{tabular*}{\textwidth}{@{\extracolsep{\fill}}lrcccccccc@{}}
        \toprule
        & & \multicolumn{4}{c}{ArcFace} & \multicolumn{4}{c}{MagFace} \\
        \cmidrule(lr){3-6}\cmidrule(lr){7-10}
        Dataset & $N$ & Mean$\pm$Std & $<0.7$ & $<0.6$ & $<0.5$ & Mean$\pm$Std & $<0.7$ & $<0.6$ & $<0.5$ \\
        \midrule
        CelebA-Test-Ref & 2,533 & $0.612{\pm}0.074$ & 88.04\% & 45.20\% & 5.29\% & $0.672{\pm}0.072$ & 65.10\% & 14.57\% & 1.30\% \\
        FFHQ-Ref & 857 & $0.687{\pm}0.064$ & 56.36\% & 8.87\% & 0.35\% & $0.722{\pm}0.067$ & 34.07\% & 4.67\% & 0.47\% \\
        \bottomrule
    \end{tabular*}}
\end{table*}

\begin{table*}[t]
    \centering
    \caption{No-reference restoration on five benchmarks; all three columns are learned perceptual-quality metrics.}
    \label{tab:realworld_blind}
    {\scriptsize\setlength{\tabcolsep}{1.4pt}\renewcommand{\arraystretch}{0.75}
    \begin{tabular*}{\textwidth}{@{\extracolsep{\fill}}llcccllccc@{}}
        \toprule
        Method & Dataset & MUSIQ $\uparrow$ & CLIP $\uparrow$ & MANIQA $\uparrow$ &
        Method & Dataset & MUSIQ $\uparrow$ & CLIP $\uparrow$ & MANIQA $\uparrow$ \\
        \midrule
\mCodeFormer & \multirow{6}{*}{LFW} & 75.484 & 0.689 & 0.527 & \mCodeFormer & \multirow{6}{*}{Wider-Test} & 73.407 & 0.699 & 0.496 \\
\mGFPGAN & & \third{75.570} & 0.676 & \second{0.551} & \mGFPGAN & & \second{74.769} & 0.700 & \second{0.550} \\
\mVQFR & & 74.901 & \second{0.725} & \third{0.543} & \mVQFR & & 72.011 & \second{0.722} & 0.514 \\
\mRFpp & & 72.251 & \third{0.702} & 0.511 & \mRFpp & & 71.518 & \third{0.717} & 0.477 \\
\mDAEFR & & \second{75.840} & 0.697 & 0.542 & \mDAEFR & & \third{74.143} & 0.697 & \third{0.520} \\
\mIConFaceOurs & & \best{76.712} & \best{0.758} & \best{0.645} & \mIConFaceOurs & & \best{75.496} & \best{0.729} & \best{0.616} \\
        \midrule
\mCodeFormer & \multirow{6}{*}{CelebChild} & \third{74.852} & 0.686 & 0.521 & \mCodeFormer & \multirow{6}{*}{CelebA-Test} & \second{75.554} & 0.671 & 0.538 \\
\mGFPGAN & & 74.822 & 0.674 & 0.530 & \mGFPGAN & & \third{75.466} & 0.672 & \second{0.568} \\
\mVQFR & & 74.459 & \second{0.711} & \second{0.542} & \mVQFR & & 74.406 & \second{0.691} & 0.552 \\
\mRFpp & & 71.690 & \third{0.702} & 0.506 & \mRFpp & & 73.914 & \third{0.689} & \third{0.553} \\
\mDAEFR & & \second{74.883} & 0.697 & \third{0.537} & \mDAEFR & & 75.251 & 0.668 & 0.545 \\
\mIConFaceOurs & & \best{75.603} & \best{0.750} & \best{0.613} & \mIConFaceOurs & & \best{75.988} & \best{0.724} & \best{0.631} \\
        \midrule
\mCodeFormer & \multirow{6}{*}{WebPhoto} & \third{74.004} & 0.692 & \third{0.503} & & & & & \\
\mGFPGAN & & \second{75.213} & \second{0.702} & \second{0.543} & & & & & \\
\mVQFR & & 71.602 & 0.690 & 0.502 & & & & & \\
\mRFpp & & 71.487 & \third{0.695} & 0.490 & & & & & \\
\mDAEFR & & 72.705 & 0.669 & 0.494 & & & & & \\
\mIConFaceOurs & & \best{75.795} & \best{0.719} & \best{0.593} & & & & & \\
        \bottomrule
    \end{tabular*}}
\end{table*}

\begin{table*}[t]
    \centering
    \caption{Component ablations on FFHQ-Ref Moderate and Severe. All retain dense concat; RefAvg uses all protocol references.}
    \label{tab:iconface_ablation_ref}
    {\scriptsize\setlength{\tabcolsep}{1.8pt}\renewcommand{\arraystretch}{0.72}
    \begin{tabular}{@{}llccccc@{}}
        \toprule
        Dataset & Variant & RefAvg$_{\mathrm{ArcFace}}$ $\uparrow$ & RefAvg$_{\mathrm{MagFace}}$ $\uparrow$ & MUSIQ $\uparrow$ & CLIP-IQA $\uparrow$ & MANIQA $\uparrow$ \\
        \midrule
        \multirow{5}{*}{FFHQ-Ref Moderate}
        & Concat & 0.644 & 0.675 & 75.917 & 0.724 & 0.599 \\
        & Concat + Struct & 0.645 & 0.675 & 75.980 & \third{0.733} & 0.606 \\
        & Concat + ID & \third{0.663} & \third{0.695} & \third{76.028} & 0.716 & \third{0.609} \\
        & ID + Struct (1R) & \second{0.674} & \second{0.702} & \second{76.135} & \second{0.736} & \second{0.611} \\
        & Full (2R) & \best{0.688} & \best{0.717} & \best{76.220} & \best{0.738} & \best{0.613} \\
        \midrule
        \multirow{5}{*}{FFHQ-Ref Severe}
        & Concat & 0.602 & 0.632 & 75.205 & 0.717 & 0.595 \\
        & Concat + Struct & 0.612 & 0.642 & 75.316 & \third{0.723} & \second{0.597} \\
        & Concat + ID & \third{0.634} & \third{0.672} & \third{75.428} & 0.721 & \third{0.596} \\
        & ID + Struct (1R) & \second{0.663} & \second{0.693} & \second{75.485} & \second{0.725} & \second{0.597} \\
        & Full (2R) & \best{0.682} & \best{0.713} & \best{75.679} & \best{0.726} & \best{0.600} \\
        \bottomrule
    \end{tabular}}
\end{table*}

\section{Experiments}

\subsection{Experimental Setup}
We train for 12 epochs on the official identity-based FFHQ-Ref split~\cite{karras2019stylegan,hsiao2024refldm} at $512{\times}512$, using online Real-ESRGAN/BSRGAN degradations~\cite{wang2021realesrgan,zhang2021bsrgan}. FFHQ-Ref val is used only qualitatively. Results use 12 steps, guidance 4.0, and seed 42; other settings are supplementary.

Using the released target--reference mappings~\cite{hsiao2024refldm}, reference-aware tests cover CelebA-Test-Ref (2,533)~\cite{liu2015celeba}, FFHQ-Ref Moderate (857), and Severe (857), against DMDNet, ReF-LDM, InstantRestore, FaceMe, and RefSTAR~\citep{li2022dmdnet,hsiao2024refldm,zhang2024instantrestore,liu2025faceme,yin2026refstar}. Blind tests cover CelebA-Test (3,000), LFW (1,711)~\cite{huang2007lfw}, CelebChild (360), WebPhoto (407), and Wider-Test (970), against CodeFormer, VQFR, GFP-GAN, RestoreFormer++, and DAEFR~\citep{zhou2022codeformer,gu2022vqfr,wang2021gfpgan,wang2023restoreformerpp,tsai2024daefr}.

We evaluate reference compatibility with RefAvg under independent ArcFace/MagFace and learned perceptual quality with MUSIQ, CLIP-IQA, and MANIQA~\cite{ke2021musiq,wang2023clipiqa,yang2022maniqa}. The supplement additionally reports paired-GT identity, paired-target reconstruction metrics~\cite{wang2004ssim,zhang2018lpips}, and target-state diagnostics for pose, expression, and semantic parsing. All IConFace results use fixed model parameters and inference settings; only reference availability changes, without mode-specific weights or test-time optimization. Baselines follow their intended protocols.

\subsection{Why Reference Similarity?}
Reference-aware restoration cannot be evaluated solely by agreement with one paired target. Same-identity references differ in pose, expression, illumination, and incidental appearance, so even clean references need not closely match the paired GT. Table~\ref{tab:main_ref_gt_gap} quantifies this cross-capture gap. On CelebA-Test-Ref, sample-level GT--reference means are $0.612/0.672$ for ArcFace/MagFace, and 88.04\%/65.10\% of samples fall below 0.7. On FFHQ-Ref, the corresponding means are $0.687/0.722$, with 56.36\%/34.07\% below 0.7. Thus, even a faithful reconstruction of one target capture need not approach unit similarity to the protocol references.

This cross-capture gap motivates a separate reference-based measure. RefAvg computes output-to-reference cosine similarities for every reference available under a method's protocol, averages the similarities within each sample, and then averages across the dataset. It neither averages reference embeddings beforehand nor selects the most favorable reference, reducing dependence on any single capture.

Training uses AdaFace for identity supervision, whereas evaluation uses independent ArcFace and MagFace recognizers. The clean GT--reference distribution in Table~\ref{tab:main_ref_gt_gap} calibrates the scale of RefAvg rather than defining a restoration upper bound. Accordingly, RefAvg measures compatibility with supplied cross-capture identity evidence, while paired-GT fidelity measures agreement with the target capture; supplementary target-state diagnostics provide additional evidence of target-state retention.

\subsection{Reference-Aware Restoration}

Table~\ref{tab:main_reference_results} follows each method's official reference use and evaluates RefAvg against the references actually supplied to that method. The supplement additionally reports a common-Ref$_1$ controlled comparison in which every method receives and is evaluated against the same protocol reference.

IConFace achieves the highest RefAvg$_{\mathrm{ArcFace}}$ and RefAvg$_{\mathrm{MagFace}}$ on all three benchmarks. Under Severe degradation, its margins over the strongest baseline for each recognizer are $0.094$ and $0.085$, substantially larger than on the easier splits. These margins support the task motivation: external identity evidence becomes more valuable as the degraded observation loses person-specific information.

IConFace also achieves the highest CLIP-IQA and MANIQA on all three benchmarks and the highest MUSIQ on CelebA-Test-Ref and FFHQ-Ref Moderate. Under Severe degradation, it combines the strongest RefAvg with the highest CLIP-IQA and MANIQA. Its RefAvg scores fall near the calibrated GT--reference cross-capture range in Table~\ref{tab:main_ref_gt_gap}.

Figure~\ref{fig:qual_ref_compact} provides qualitative comparisons; on Severe, supplementary pose/expression/parsing diagnostics yield target-preference rates of 95.0/73.0/94.9\%.

\subsection{Persistent Localized Identity Detail Preservation}

RefAvg measures global identity compatibility but cannot determine whether a specific localized trait survives restoration. From 235 high-recall candidates, cross-image human audit retains 167 cases whose primary trait is supported by both the paired GT and at least one same-identity reference. Each output is evaluated against the paired GT using the audited trait description; the supplement details the complete protocol.

As shown in Fig.~\ref{fig:main_pid_preservation_rate}, IConFace preserves 160 of 167 audited traits (95.8\%), achieving the highest observed preservation rate among all evaluated methods, compared with 92.2\% for the next-best method. Figure~\ref{fig:main_visible_feature_examples} illustrates a representative case in which IConFace preserves all three moles along the cheek at their corresponding locations, whereas competing methods omit or weaken part of this localized pattern.

\subsection{No-Reference Restoration}
Table~\ref{tab:realworld_blind} compares IConFace with five established blind face restorers across five benchmarks. IConFace achieves leading learned perceptual quality across these benchmarks. Figure~\ref{fig:qual_noref_compact} shows sharp, coherent restorations across diverse ages, poses, occlusions, illumination, and degradation conditions. In particular, IConFace recovers clearer eyes and facial-component details, preserves visible occlusion boundaries and surrounding facial structure, and reduces the over-smoothed or locally distorted appearance observed in several competing outputs. The supplement additionally reports paired-target reconstruction on CelebA-Test.

\subsection{Ablation Studies}
We evaluate five architecture variants under the same training protocol on FFHQ-Ref Moderate and Severe. \textit{Concat} is the dense-token baseline; \textit{Concat + Struct} and \textit{Concat + ID} assess the two conditioning pathways individually, \textit{ID + Struct (1R)} combines them with one degraded-memory route, and \textit{Full (2R)} uses the proposed two-route memory.

Table~\ref{tab:iconface_ablation_ref} reveals a clear division of roles between the two pathways. Degraded-structure reinforcement primarily improves learned perceptual quality, whereas identity guidance produces larger gains in reference compatibility, particularly under Severe degradation. Enabling both pathways improves identity compatibility and perceptual quality together. The comparison between \textit{ID + Struct (1R)} and \textit{Full (2R)} further supports the two-route design: separate full-field and local-residual memories yield the strongest value for every reported metric on both splits. Figure~\ref{fig:iconface_ablation_ref_compact} shows the same progression qualitatively. Structure reinforcement stabilizes facial geometry, identity guidance recovers a more reference-compatible appearance, and the full model combines clearer local features with target-aligned pose and structure, with the effect most apparent under Severe degradation. Additional examples appear in the supplement.

\section{Conclusion}
We propose IConFace, retaining degraded and optional reference observations as dense tokens with compact identity guidance and full-field/local-residual structure memory. Across three reference-aware benchmarks, it achieves the strongest reference compatibility, particularly under severe degradation, and the highest observed rate on a human-audited localized-detail benchmark. It also delivers leading learned perceptual quality on five blind-restoration benchmarks, demonstrating strong performance in both settings.

\clearpage
\setlength{\bibsep}{0pt}
\bibliography{ref}

\expandafter\let\csname ablimgw\endcsname\relax
\expandafter\let\csname best\endcsname\relax
\expandafter\let\csname datasetheader\endcsname\relax
\expandafter\let\csname fingerroot\endcsname\relax
\expandafter\let\csname fingersize\endcsname\relax
\expandafter\let\csname mCodeFormer\endcsname\relax
\expandafter\let\csname mDAEFR\endcsname\relax
\expandafter\let\csname mDMDNet\endcsname\relax
\expandafter\let\csname mFaceMe\endcsname\relax
\expandafter\let\csname mGFPGAN\endcsname\relax
\expandafter\let\csname mInstantRestore\endcsname\relax
\expandafter\let\csname mRFpp\endcsname\relax
\expandafter\let\csname mReFLDM\endcsname\relax
\expandafter\let\csname mRefSTAR\endcsname\relax
\expandafter\let\csname mVQFR\endcsname\relax
\expandafter\let\csname methodvenue\endcsname\relax
\expandafter\let\csname qfigsetup\endcsname\relax
\expandafter\let\csname qhead\endcsname\relax
\expandafter\let\csname qheadspaced\endcsname\relax
\expandafter\let\csname qimg\endcsname\relax
\expandafter\let\csname qimgarcscorepath\endcsname\relax
\expandafter\let\csname qimgpath\endcsname\relax
\expandafter\let\csname qrowsep\endcsname\relax
\expandafter\let\csname qrowsepcompact\endcsname\relax
\expandafter\let\csname qslot\endcsname\relax
\expandafter\let\csname second\endcsname\relax
\expandafter\let\csname teaserimg\endcsname\relax
\expandafter\let\csname third\endcsname\relax

\definecolor{bestcolor}{RGB}{196,48,43}
\definecolor{secondcolor}{RGB}{54,102,178}
\definecolor{thirdcolor}{RGB}{52,145,78}
\newcommand{\best}[1]{\textbf{\textcolor{bestcolor}{#1}}}
\newcommand{\second}[1]{\textbf{\textcolor{secondcolor}{#1}}}
\newcommand{\third}[1]{\textbf{\textcolor{thirdcolor}{#1}}}
\newcommand{\fingersize}{512}
\newcommand{\fingerroot}{fingers/iconface_qualitative/finger_\fingersize}
\newcommand{\qcolsep}{0.0045\textwidth}
\newcommand{\qslot}[2]{\begin{minipage}[t]{#1}\vspace{0pt}\centering #2\end{minipage}}
\newcommand{\qhead}[2]{\qslot{#1}{{\scriptsize #2}}}
\newcommand{\qheadspaced}[2]{\qslot{#1}{{\scriptsize #2}\par\vspace{1.8pt}}}
\newcommand{\qhardhead}[2]{\qslot{#1}{{\fontsize{6.1}{6.4}\selectfont #2}}}
\newcommand{\qimg}[2]{\qslot{#1}{\includegraphics[width=\linewidth]{\fingerroot/#2}}}
\newcommand{\qimgpath}[2]{\qslot{#1}{\includegraphics[width=\linewidth]{#2}}}
\newcommand{\qimgarcscorepath}[3]{%
\begin{minipage}[t]{#1}\centering
\vbox{
  \hbox to \linewidth{\hfil\includegraphics[width=\linewidth]{#2}\hfil}
  \kern-10.0pt
  \hbox to \linewidth{\hfil{\begingroup\setlength{\fboxsep}{0.2pt}\colorbox{white}{\fontsize{5.6}{5.6}\selectfont Arc~#3}\endgroup}\kern0.4pt}
}
\end{minipage}}
\newcommand{\qdiaglabel}[3]{\qslot{#1}{{\fontsize{5.7}{5.9}\selectfont\textbf{#2}\\[-0.6pt]\texttt{#3}}}}
\newcommand{\qblank}[1]{\qslot{#1}{\rule{\linewidth}{0pt}\rule{0pt}{\linewidth}}}
\newcommand{\qtodo}[2]{\qslot{#1}{\fcolorbox{black!30}{black!5}{\parbox[c][#1][c]{#1}{\centering\scriptsize #2}}}}
\newcommand{\qplaceholder}[1]{\qslot{#1}{\fcolorbox{black!30}{black!5}{\parbox[c][#1][c]{#1}{\centering}}}}
\newcommand{\qlabeled}[3]{\qslot{#1}{\includegraphics[width=\linewidth]{\fingerroot/#2}}}
\newcommand{\qlabeledpath}[3]{\qslot{#1}{\includegraphics[width=\linewidth]{#2}}}
\newcommand{\qrefoverlapone}[1]{\makebox[\linewidth][c]{\includegraphics[width=0.78\linewidth]{#1}\rule{0pt}{0.80\linewidth}}}
\newcommand{\qrefoverlaptwo}[2]{\makebox[\linewidth][c]{\makebox[0pt][l]{\hspace*{0.08\linewidth}\raisebox{0.10\linewidth}[0pt][0pt]{\includegraphics[width=0.56\linewidth]{#1}}}\makebox[0pt][l]{\hspace*{0.34\linewidth}\raisebox{0.00\linewidth}[0pt][0pt]{\includegraphics[width=0.56\linewidth]{#2}}}\rule{0pt}{0.78\linewidth}}}
\newcommand{\qrefoverlapthree}[3]{\makebox[\linewidth][c]{\makebox[0pt][l]{\hspace*{0.18\linewidth}\raisebox{0.16\linewidth}[0pt][0pt]{\includegraphics[width=0.44\linewidth]{#1}}}\makebox[0pt][l]{\hspace*{0.03\linewidth}\raisebox{0.00\linewidth}[0pt][0pt]{\includegraphics[width=0.44\linewidth]{#2}}}\makebox[0pt][l]{\hspace*{0.35\linewidth}\raisebox{0.02\linewidth}[0pt][0pt]{\includegraphics[width=0.44\linewidth]{#3}}}\rule{0pt}{0.78\linewidth}}}
\newcommand{\qrefone}[3]{\qslot{#1}{\qrefoverlapone{\fingerroot/#2}}}
\newcommand{\qreftwo}[4]{\qslot{#1}{\qrefoverlaptwo{\fingerroot/#2}{\fingerroot/#3}}}
\newcommand{\qrefthree}[5]{\qslot{#1}{\qrefoverlapthree{\fingerroot/#2}{\fingerroot/#3}{\fingerroot/#4}}}
\newcommand{\qrefonepath}[3]{\qslot{#1}{\qrefoverlapone{#2}}}
\newcommand{\qreftwopath}[4]{\qslot{#1}{\qrefoverlaptwo{#2}{#3}}}
\newcommand{\qrefthreepath}[5]{\qslot{#1}{\qrefoverlapthree{#2}{#3}{#4}}}
\newcommand{\refimgw}{0.112\textwidth}
\newcommand{\refstackw}{\refimgw}
\newcommand{\hardimgw}{0.0680\textwidth}
\newcommand{\hardrefw}{\hardimgw}
\newcommand{\blindpairw}{0.112\textwidth}
\newcommand{\blindrealw}{0.127\textwidth}
\newcommand{\ablimgw}{0.100\textwidth}
\newcommand{\diaglblw}{0.070\textwidth}
\newcommand{\diagimgw}{0.102\textwidth}
\newcommand{\hardcasefigroot}{fingers/iconface_hard20}
\newcommand{\datasetheader}[1]{{\small\textbf{#1}}\par\vspace{0.5pt}}
\newcommand{\qfigsetup}{\setlength{\tabcolsep}{1.3pt}\renewcommand{\arraystretch}{0.94}}
\newlength{\qrowsep}
\setlength{\qrowsep}{1.3pt}
\newcommand{\diagimgwcompact}{0.100\textwidth}
\newcommand{\supprefw}{0.101\textwidth}
\newcommand{\suppblindw}{0.1065\textwidth}
\newcommand{\suppablw}{0.102\textwidth}
\newcommand{\suppcaseid}[2]{\multicolumn{#1}{c}{{\fontsize{5.9}{5.9}\selectfont #2}}\\[-1.8pt]}
\newcommand{\suppblindcaseid}[2]{\noalign{\vskip 0pt}\multicolumn{#1}{c}{\raisebox{-0.3pt}{{\fontsize{5.9}{5.9}\selectfont #2}}}\\[-2.4pt]}
\newcommand{\datasetheadercompact}[1]{{\small\textbf{#1}}\par\vspace{0.5pt}}
\newcommand{\qfigsetupcompact}{\setlength{\tabcolsep}{0.40pt}\renewcommand{\arraystretch}{0.88}}
\newcommand{\qfigsetupblindcompact}{\setlength{\tabcolsep}{0.35pt}\renewcommand{\arraystretch}{0.80}}
\newlength{\qrowsepcompact}
\setlength{\qrowsepcompact}{0pt}
\newcommand{\qheadcompact}[2]{\qslot{#1}{{\scriptsize #2}\par\vspace{1.0pt}}}
\newcommand{\methodvenue}[2]{\mbox{#1 {\fontsize{5.7}{5.7}\selectfont (#2)}}}
\newcommand{\mDMDNet}{\methodvenue{DMDNet}{TPAMI'22}}
\newcommand{\mReFLDM}{\methodvenue{ReF-LDM}{NeurIPS'24}}
\newcommand{\mInstantRestore}{\methodvenue{InstantRestore}{SIGGRAPH'25}}
\newcommand{\mFaceMe}{\methodvenue{FaceMe}{AAAI'25}}
\newcommand{\mRefSTAR}{\methodvenue{RefSTAR}{AAAI'26}}
\newcommand{\mCodeFormer}{\methodvenue{CodeFormer}{NeurIPS'22}}
\newcommand{\mGFPGAN}{\methodvenue{GFP-GAN}{CVPR'21}}
\newcommand{\mVQFR}{\methodvenue{VQFR}{ECCV'22}}
\newcommand{\mRFpp}{\methodvenue{RF++}{TPAMI'23}}
\newcommand{\mDAEFR}{\methodvenue{DAEFR}{ICLR'24}}
\newcommand{\supptablefont}{\scriptsize}
\newcommand{\harddatasetheader}[1]{{\footnotesize\textbf{#1}}\par\vspace{0.5pt}}
\newcommand{\qhardfigsetup}{\setlength{\tabcolsep}{0.9pt}\renewcommand{\arraystretch}{0.82}}
\newcommand{\teaserbox}[3]{\fcolorbox{black!35}{black!5}{\parbox[c][#2][c]{#1}{\centering\scriptsize #3}}}
\newcommand{\teaserimg}[3]{\parbox[c][#2][c]{#1}{\centering\includegraphics[width=#1,height=#2]{#3}}}

\renewcommand{\topfraction}{0.95}
\renewcommand{\bottomfraction}{0.90}
\renewcommand{\textfraction}{0.03}
\renewcommand{\floatpagefraction}{0.85}
\renewcommand{\dbltopfraction}{0.95}
\renewcommand{\dblfloatpagefraction}{0.85}
\setcounter{topnumber}{5}
\setcounter{bottomnumber}{5}
\setcounter{totalnumber}{10}
\setcounter{dbltopnumber}{4}
\setlength{\floatsep}{4pt plus 1pt minus 1pt}
\setlength{\textfloatsep}{6pt plus 1pt minus 1pt}
\setlength{\dblfloatsep}{4pt plus 1pt minus 1pt}
\setlength{\dbltextfloatsep}{6pt plus 1pt minus 1pt}

\newcommand{\supprefrow}[3]{%
    \suppcaseid{9}{#1: case #2}
    \qimgpath{\supprefw}{\hardcasefigroot/#1/#2__ref1.#3} &
    \qimgpath{\supprefw}{\hardcasefigroot/#1/#2__deg.jpg} &
    \qimgpath{\supprefw}{\hardcasefigroot/#1/#2__dmdnet.jpg} &
    \qimgpath{\supprefw}{\hardcasefigroot/#1/#2__refldm.jpg} &
    \qimgpath{\supprefw}{\hardcasefigroot/#1/#2__instantrestore.jpg} &
    \qimgpath{\supprefw}{\hardcasefigroot/#1/#2__faceme.jpg} &
    \qimgpath{\supprefw}{\hardcasefigroot/#1/#2__refstar.jpg} &
    \qimgpath{\supprefw}{\hardcasefigroot/#1/#2__ours.jpg} &
    \qimgpath{\supprefw}{\hardcasefigroot/#1/#2__gt.jpg}\\[\qrowsepcompact]}
\newcommand{\suppblindrow}[3]{%
    \suppblindcaseid{7}{#1: case \texttt{\detokenize{#3}}}
    \qimgpath{\suppblindw}{#2/#3__deg.jpg} &
    \qimgpath{\suppblindw}{#2/#3__codeformer.jpg} &
    \qimgpath{\suppblindw}{#2/#3__gfpgan.jpg} &
    \qimgpath{\suppblindw}{#2/#3__vqfr.jpg} &
    \qimgpath{\suppblindw}{#2/#3__restoreformerpp.jpg} &
    \qimgpath{\suppblindw}{#2/#3__daefr.jpg} &
    \qimgpath{\suppblindw}{#2/#3__ours.jpg}\\[\qrowsepcompact]}
\newcommand{\suppablscoredrow}[9]{%
    \suppcaseid{8}{case \texttt{\detokenize{#2}}}
    \qimgarcscorepath{\suppablw}{#1/#2__ref1.jpg}{1.000} &
    \qimgarcscorepath{\suppablw}{#1/#2__deg.jpg}{#3} &
    \qimgarcscorepath{\suppablw}{#1/#2__concat.jpg}{#4} &
    \qimgarcscorepath{\suppablw}{#1/#2__struct.jpg}{#5} &
    \qimgarcscorepath{\suppablw}{#1/#2__id.jpg}{#6} &
    \qimgarcscorepath{\suppablw}{#1/#2__1r.jpg}{#7} &
    \qimgarcscorepath{\suppablw}{#1/#2__full.jpg}{#8} &
    \qimgarcscorepath{\suppablw}{#1/#2__gt.jpg}{#9}\\[\qrowsepcompact]}

\clearpage
\setcounter{figure}{0}
\setcounter{table}{0}
\setcounter{equation}{0}
\setcounter{section}{0}
\twocolumn[
\begin{center}
{\LARGE\bfseries Supplementary Material for\\
IConFace: Fine-Grained Identity Conditioning for\\
Reference-Aware Face Restoration\par}
\vspace{0.8em}
{\large\bfseries Axi Niu$^*$, Jinyang Zhang$^*$, Senyan Qing\par}
\vspace{0.35em}
Northwestern Polytechnical University\\
nax@nwpu.edu.cn \quad zhangjinyang@mail.nwpu.edu.cn \quad qingsenyan@nwpu.edu.cn\\
Homepage: \url{https://cosmicrealm.github.io/IConFace/}\\
\end{center}
\vspace{0.8em}
]
\begingroup
\renewcommand{\thefootnote}{*}
\footnotetext{These authors contributed equally.}
\endgroup

\appendix
\section{Supplementary Contents}
This supplement follows the experimental order of the main paper and separates reproducibility details, quantitative controls, and visual evidence. The compact map below states the role of each module.

{\small
\begin{itemize}[leftmargin=1.2em,itemsep=1.5pt,parsep=0pt,topsep=2pt]
    \item \textbf{Implementation and Evaluation Protocols} supports the Method and Experimental Setup with training, conditioning, token-packing, and metric definitions.
    \item \textbf{Reference-Aware Evaluation} supports the reference-aware results through a common-Ref$_1$ control, reference-count analysis, paired-target fidelity, and IConFace target-versus-reference state alignment.
    \item \textbf{Localized Identity-Detail Evaluation} documents benchmark construction, the human-verified protocol, and preservation results.
    \item \textbf{No-Reference Evaluation} complements the blind-restoration results with paired-target reconstruction evidence.
    \item \textbf{Ablation Analysis} extends the main-paper ablation with qualitative component comparisons on both FFHQ-Ref splits.
\end{itemize}}
\section{Implementation and Evaluation Protocols}
\label{sec:supp_implementation}

\subsection{Training and Inference}
\paragraph{Training data.}
FFHQ-Ref is the reference-aware benchmark released with ReF-LDM. We directly use its official identity-based train/validation/test partitions and target--reference mappings without modifying the identity partition or reference assignments. We use the training split for model fitting and reserve the validation split for preview and qualitative validation. Each clean target is resized and center-cropped to $512{\times}512$, then degraded online by selecting BSRGAN- or Real-ESRGAN-style synthesis with equal probability. The BSRGAN branch randomizes the order of isotropic or anisotropic blur, staged downsampling, Gaussian noise, and JPEG compression. The Real-ESRGAN branch applies a two-stage composition of mixed Gaussian, generalized, plateau, or sinc kernels, stochastic resizing, Gaussian or Poisson noise, and JPEG compression. The resulting LQ image is bicubically resized back to $512{\times}512$. Same-identity references are sampled independently and receive standard spatial preprocessing but no synthetic degradation.

\paragraph{Optimization and inference.}
IConFace uses the FLUX.2-klein-base-4B restoration backbone and adds rank 16 LoRA adapters together with the identity and degraded-structure modules described in the main paper. LoRA is applied to image-attention QKV/projection and image-MLP linear layers. The final model has about 139.8M trainable parameters. Training uses $512{\times}512$ crops on four GPUs in bf16, with per-device batch size 1 and gradient accumulation 4, giving an effective global batch size of 16. We train the final model for 12 epochs. AdamW uses weight decay $10^{-4}$, 100 warmup steps, and cosine decay. The base learning rate is $10^{-5}$; LoRA uses $5{\times}10^{-7}$, the identity projector uses $3{\times}10^{-5}$, and other restoration parameters use $10^{-5}$. Degradation strength is sampled from 0--16 with bucket probabilities 0.5, 0.3, and 0.2 for ranges 0--3, 4--8, and 9--16. All reported results use 12 sampling steps, guidance scale 4.0, base seed 42, and 512 resolution.

\subsection{Reference Sampling and Conditioning}
Training samples $N=0/1/2/3$ references with probabilities 30/30/20/20, respectively. When fewer references are available, the actual number is used without duplication. The identity pathway uses AdaFace IR50 embeddings with norm-weighted multi-reference aggregation at temperature 1.0. Reference and degraded visual tokens remain in the concat sequence, so the identity pathway is an additional reliability bias rather than the sole carrier of reference evidence.

The degraded-structure pathway injects a rank 16 low-rank input residual with strength 1.0 and supplies 256 pooled memory tokens to all 5 double-stream and 20 single-stream blocks through rank 16 K/V projectors. Its 171 full-field and 85 local-residual tokens provide complementary degraded-side context. The local residual is formed by subtracting $5{\times}5$ average-pooled features, emphasizing spatially localized deviations around the pooled full-field representation.

\subsection{Token Packing and Source-Aware 3D RoPE}
All image and text conditions use the four-axis FLUX.2 rotary position layout $(t,h,w,l)$. Scene and degraded latents share the same spatial grid and are independently packed from $(B,C,H,W)$ to $(B,HW,C)$ without interpolation or spatial resampling. Scene tokens use ids $(0,h,w,0)$, while text tokens use ids $(0,0,0,l)$. The degraded image condition is assigned a separate temporal group with $t=2$. Reference images are packed as grouped visual tokens with $t=10+r$ for reference index $r$, which lets the transformer distinguish the generated scene, degraded observation, and each reference without changing the spatial $(h,w)$ axes. For a token $i$ with position id $p_i=(t_i,h_i,w_i,l_i)$, the query and key vectors are split over axes $a\in\{t,h,w,l\}$ with dimensions $d_a=[32,32,32,32]$. For each two-channel pair $m$ on axis $a$, the RoPE computation is
\begin{subequations}
\begin{align}
\omega_{a,m} &= \theta^{-2m/d_a},\qquad \theta=2000,\\
R(\rho,\omega) &=
\left[\begin{smallmatrix}
\cos(\rho\omega) & -\sin(\rho\omega)\\
\sin(\rho\omega) & \cos(\rho\omega)
\end{smallmatrix}\right],\\
\tilde{q}_{i,a,m} &= R(p_i^a,\omega_{a,m})q_{i,a,m},\\
\tilde{k}_{j,a,m} &= R(p_j^a,\omega_{a,m})k_{j,a,m},\\
\mathrm{Attn}_{ij} &\propto
\exp\!\left(\frac{\tilde{q}_i^\top \tilde{k}_j}{\sqrt{d}}\right).
\end{align}
\end{subequations}
This construction preserves the ordinary spatial axes while using the temporal axis to mark scene, degraded, and reference token groups.

\subsection{Evaluation Protocols and Metrics}
Reference-aware evaluation follows the official CelebA-Test-Ref and FFHQ-Ref test sample IDs, paired targets, and reference mappings released with ReF-LDM. The two FFHQ-Ref splits share clean targets and references but use different degradation strengths. No-reference evaluation uses paired CelebA-Test and the real-world LFW, CelebChild, WebPhoto, and Wider-Test sets. The headline reference-aware comparison follows each method's officially supported reference configuration, and each output is evaluated only against the method-specific reference set supplied under that configuration. We additionally conduct a common-Ref$_1$ controlled comparison in which every method receives and is evaluated against the same protocol reference.

For each reference-aware sample, let $\hat{I}_i^{(m)}$ be the output of method $m$, $G_i$ be the paired GT target, and $\mathcal{R}_i^{(m)}$ be the method-specific reference set supplied under the evaluated configuration. RefAvg$_{\mathrm{ArcFace}}$ and RefAvg$_{\mathrm{MagFace}}$ are dataset averages of per-sample mean cosine similarities:
Define the encoder-specific cosine similarity as
\begin{equation}
s_E(X,Y)=\frac{E(X)^\top E(Y)}{\|E(X)\|_2\,\|E(Y)\|_2}.
\end{equation}
Then
\begin{equation}
\begin{aligned}
\mathrm{RefAvgMetric}(m)
&=\frac{1}{N}\sum_{i=1}^{N}\frac{1}{|\mathcal{R}_i^{(m)}|}\\
&\quad\sum_{R\in\mathcal{R}_i^{(m)}}
s_E\!\left(\hat{I}_i^{(m)},R\right).
\end{aligned}
\end{equation}
where $E$ is the corresponding frozen ArcFace or MagFace encoder. The average is taken over similarities, not by first averaging reference embeddings. GT$_{\mathrm{ArcFace}}$ and GT$_{\mathrm{MagFace}}$ compare $\hat{I}_i^{(m)}$ with $G_i$. ArcFace and MagFace are independent of the training-time AdaFace anchor. PSNR, SSIM, and LPIPS serve as paired-target reconstruction metrics. MUSIQ, CLIP-IQA, and MANIQA measure learned perceptual quality without requiring a paired target.

For target-state diagnostics, all inputs are resized to $512{\times}512$. InsightFace \texttt{buffalo\_l} selects the largest detected face and extracts 68 facial landmarks and three Euler pose angles. Detection uses a $640{\times}640$ input and a threshold of 0.5; failed cases are retried with 25\% reflected padding. Pose error is the mean absolute angular difference over the three pose components. The expression proxy comprises mouth opening, mean eye opening, and smile width, each normalized by the corresponding image's interocular distance; expression error averages their absolute differences. Semantic parsing uses the 19-class ParseNet and reports mean IoU over seven grouped regions: skin, brows, eyes, nose, mouth/lips, hair, and face foreground. Failures are excluded only from the affected metric denominator and recorded separately; all IConFace samples reported in the target-state analysis are valid for all three diagnostics.

\FloatBarrier
\section{Reference-Aware Evaluation}
\label{sec:supp_reference_aware}
The four analyses below extend the main-paper reference-aware evaluation. A common-Ref$_1$ control matches supplied identity evidence across methods, the reference-count diagnostic measures sensitivity to reference availability, paired-target metrics evaluate agreement with one target capture, and target-versus-reference margins quantify retention of the target capture state.

\subsection{Common-Ref$_1$ Controlled Comparison}
\label{sec:supp_common_ref1}

The main comparison follows each method's officially supported reference configuration. To control the supplied identity evidence, we additionally evaluate every method using the same protocol Ref$_1$. Each method receives this image through the reference interface of its official implementation, and no other same-identity reference is supplied. Under this setting, the method-specific reference set becomes $\mathcal{R}_i^{(m)}=\{R_{i,1}\}$ for every method, so RefAvg reduces to the output-to-Ref$_1$ cosine similarity.

For identity encoder $E$, the controlled similarity is
\begin{equation}
\mathrm{Ref1Sim}_{E}(m)
=
\frac{1}{N}
\sum_{i=1}^{N}
s_E\!\left(\hat{I}_i^{(m,1)},R_{i,1}\right),
\label{eq:supp_ref1sim}
\end{equation}
where $\hat{I}_i^{(m,1)}$ is the output of method $m$ conditioned on the common protocol Ref$_1$.
For DMDNet, whose specific dictionary interface expects two exemplar tensors, the wrapper duplicates Ref$_1$ internally rather than supplying a second distinct identity image.

\begin{table*}[!t]
    \centering
    \caption{Controlled reference-aware comparison using the common protocol Ref$_1$. Every method receives the same Ref$_1$ through the reference interface of its official implementation. Ref$_1$-Sim is the cosine similarity between the restored output and that reference; perceptual metrics are computed from the same restored outputs.}
    \label{tab:supp_common_ref1}
    {\supptablefont\setlength{\tabcolsep}{2.2pt}\renewcommand{\arraystretch}{0.92}
    \begin{tabular}{@{}llccccc@{}}
        \toprule
        Method & Dataset & Ref$_1$-Sim$_{\mathrm{ArcFace}}$ $\uparrow$ & Ref$_1$-Sim$_{\mathrm{MagFace}}$ $\uparrow$ & MUSIQ $\uparrow$ & CLIP-IQA $\uparrow$ & MANIQA $\uparrow$ \\
        \midrule
        \mDMDNet & \multirow{6}{*}{\shortstack{CelebA-Test\\Ref}} & 0.469 & 0.525 & \third{72.780} & 0.625 & 0.494 \\
        \mReFLDM & & \third{0.569} & \third{0.620} & \second{73.662} & \second{0.660} & \second{0.513} \\
        \mInstantRestore & & 0.545 & 0.601 & 71.046 & 0.551 & 0.490 \\
        \mFaceMe & & 0.548 & 0.601 & 70.666 & 0.580 & 0.481 \\
        \mRefSTAR & & \second{0.594} & \second{0.652} & 71.633 & \third{0.631} & \third{0.509} \\
        IConFace & & \best{0.693} & \best{0.725} & \best{76.065} & \best{0.721} & \best{0.629} \\
        \midrule
        \mDMDNet & \multirow{6}{*}{\shortstack{FFHQ-Ref\\Moderate}} & 0.566 & 0.610 & 72.010 & 0.631 & 0.476 \\
        \mReFLDM & & 0.630 & 0.667 & \second{74.303} & \second{0.692} & \second{0.521} \\
        \mInstantRestore & & 0.610 & 0.652 & 68.946 & 0.563 & 0.458 \\
        \mFaceMe & & \third{0.643} & \third{0.682} & 72.249 & 0.608 & 0.491 \\
        \mRefSTAR & & \second{0.648} & \second{0.686} & \third{72.890} & \third{0.646} & \third{0.501} \\
        IConFace & & \best{0.730} & \best{0.753} & \best{76.191} & \best{0.727} & \best{0.604} \\
        \midrule
        \mDMDNet & \multirow{6}{*}{\shortstack{FFHQ-Ref\\Severe}} & 0.135 & 0.174 & 62.077 & 0.540 & 0.353 \\
        \mReFLDM & & \third{0.556} & \third{0.588} & \second{74.924} & \second{0.703} & \second{0.527} \\
        \mInstantRestore & & 0.471 & 0.516 & 69.667 & 0.575 & 0.462 \\
        \mFaceMe & & 0.454 & 0.506 & 71.903 & 0.609 & 0.494 \\
        \mRefSTAR & & \second{0.580} & \second{0.628} & \third{73.012} & \third{0.645} & \third{0.512} \\
        IConFace & & \best{0.742} & \best{0.762} & \best{75.548} & \best{0.708} & \best{0.586} \\
        \bottomrule
    \end{tabular}}
\end{table*}

\begin{table*}[!t]
    \centering
    \caption{Reference-count diagnostic on fixed three-reference subsets: 1,921 CelebA-Test-Ref samples and 841 samples from each FFHQ-Ref split, with identical sample IDs used for all values of $N$. GT columns measure paired-target identity. Ref$_3$ is held out for $N\leq2$ and supplied at $N=3$.}
    \label{tab:supp_reference_count}
    {\supptablefont\setlength{\tabcolsep}{4.2pt}\renewcommand{\arraystretch}{0.94}
    \begin{tabular}{@{}llcccc@{}}
        \toprule
        Dataset & $N$ & GT$_{\mathrm{ArcFace}}$ $\uparrow$ & GT$_{\mathrm{MagFace}}$ $\uparrow$ & Ref$_3$-Sim$_{\mathrm{ArcFace}}$ $\uparrow$ & Ref$_3$-Sim$_{\mathrm{MagFace}}$ $\uparrow$ \\
        \midrule
        \multirow{4}{*}{CelebA-Test-Ref}
        & 0 & 0.545 & 0.560 & 0.340 & 0.394 \\
        & 1 & 0.685 & 0.713 & 0.530 & 0.590 \\
        & 2 & 0.723 & 0.747 & 0.556 & 0.615 \\
        & 3 & 0.739 & 0.763 & 0.626 & 0.675 \\
        \midrule
        \multirow{4}{*}{FFHQ-Ref Moderate}
        & 0 & 0.627 & 0.649 & 0.442 & 0.491 \\
        & 1 & 0.762 & 0.780 & 0.629 & 0.665 \\
        & 2 & 0.793 & 0.808 & 0.648 & 0.682 \\
        & 3 & 0.806 & 0.819 & 0.692 & 0.721 \\
        \midrule
        \multirow{4}{*}{FFHQ-Ref Severe}
        & 0 & 0.143 & 0.164 & 0.099 & 0.130 \\
        & 1 & 0.665 & 0.696 & 0.605 & 0.645 \\
        & 2 & 0.703 & 0.730 & 0.625 & 0.665 \\
        & 3 & 0.720 & 0.745 & 0.688 & 0.718 \\
        \bottomrule
    \end{tabular}}
\end{table*}

Under matched Ref$_1$ input, Table~\ref{tab:supp_common_ref1} shows that IConFace ranks first under both independent identity encoders and all three perceptual metrics on each dataset. RefSTAR is generally the closest identity competitor, whereas ReF-LDM is generally the closest perceptual-quality competitor. The identity margin is largest under Severe degradation: Ref$_1$-Sim$_{\mathrm{ArcFace}}$ is 0.742 for IConFace and 0.580 for RefSTAR.

\subsection{Reference-Count Diagnostic}
This diagnostic isolates the effect of reference availability within IConFace. On fixed samples for which all three protocol references are available, we run the same checkpoint and inference configuration with $N=0,1,2,3$, supplying the first $N$ references when $N>0$. We report paired-GT identity and similarity to Ref$_3$.

Table~\ref{tab:supp_reference_count} shows that paired-GT identity increases monotonically as references are added on all three datasets, with the largest change from $N=0$ to $N=1$ under Severe degradation, where GT$_{\mathrm{ArcFace}}$ rises from 0.143 to 0.665. Later references provide smaller additional gains, and similarity to Ref$_3$ also rises with reference availability.

\subsection{Paired-Target Fidelity}

The main reference-aware comparison and the paired-target results below follow each method's officially supported reference configuration. The preceding common-Ref$_1$ experiment separately controls the supplied reference image. The GT--reference distribution establishes the cross-capture gap between identity evidence and one target photograph, while the results below provide complementary paired-target fidelity and IConFace target-versus-reference state alignment without repeating the headline RefAvg scores.

\begin{table*}[!t]
    \centering
    \caption{Paired-target fidelity on the reference-aware benchmarks. GT ArcFace/MagFace measure identity agreement with the paired target; PSNR, SSIM, and LPIPS are paired-target reconstruction metrics. All methods follow the officially supported reference configurations used in the main comparison.}
    \label{tab:supp_paired_target_fidelity}
    {\supptablefont\setlength{\tabcolsep}{2.2pt}\renewcommand{\arraystretch}{0.92}
    \begin{tabular}{@{}llccccc@{}}
        \toprule
        Dataset & Method & GT$_{\mathrm{ArcFace}}$ $\uparrow$ & GT$_{\mathrm{MagFace}}$ $\uparrow$ & PSNR $\uparrow$ & SSIM $\uparrow$ & LPIPS $\downarrow$ \\
        \midrule
        \multirow{6}{*}{CelebA-Test-Ref}
        & \mDMDNet & 0.740 & 0.753 & \second{25.535} & \third{0.696} & \third{0.253} \\
        & \mReFLDM & \third{0.783} & \third{0.802} & 23.901 & 0.638 & 0.268 \\
        & \mInstantRestore & 0.770 & 0.788 & \third{25.400} & \best{0.702} & \best{0.219} \\
        & \mFaceMe & \second{0.801} & \second{0.812} & \best{25.947} & \second{0.698} & \third{0.253} \\
        & \mRefSTAR & \best{0.808} & \best{0.828} & 24.856 & 0.680 & \second{0.221} \\
        & IConFace & 0.726 & 0.750 & 22.327 & 0.632 & 0.277 \\
        \midrule
        \multirow{6}{*}{FFHQ-Ref Moderate}
        & \mDMDNet & 0.804 & 0.822 & \second{25.844} & \third{0.726} & 0.234 \\
        & \mReFLDM & \third{0.837} & \third{0.853} & 23.971 & 0.664 & 0.232 \\
        & \mInstantRestore & 0.812 & 0.831 & \third{25.452} & \second{0.727} & \third{0.213} \\
        & \mFaceMe & \best{0.870} & \best{0.883} & \best{26.335} & \best{0.733} & \best{0.173} \\
        & \mRefSTAR & \second{0.848} & \second{0.861} & 25.150 & 0.708 & \second{0.202} \\
        & IConFace & 0.808 & 0.820 & 22.755 & 0.671 & 0.218 \\
        \midrule
        \multirow{6}{*}{FFHQ-Ref Severe}
        & \mDMDNet & 0.182 & 0.207 & 20.111 & 0.570 & 0.449 \\
        & \mReFLDM & \second{0.654} & \second{0.682} & 19.563 & 0.566 & 0.347 \\
        & \mInstantRestore & 0.534 & 0.572 & \best{21.372} & \best{0.647} & \best{0.323} \\
        & \mFaceMe & 0.536 & 0.575 & \second{20.606} & \second{0.616} & \third{0.338} \\
        & \mRefSTAR & \third{0.632} & \third{0.671} & \third{20.404} & \third{0.610} & \second{0.328} \\
        & IConFace & \best{0.721} & \best{0.746} & 18.376 & 0.570 & 0.357 \\
        \bottomrule
    \end{tabular}}
\end{table*}

Table~\ref{tab:supp_paired_target_fidelity} shows that IConFace achieves the strongest paired-GT identity under Severe degradation, where the degraded observation contains the least recoverable identity evidence.

\subsection{Target-versus-Reference State Alignment}
Under severe degradation, stronger reliance on same-identity references is expected because the degraded observation may no longer contain sufficient person-specific evidence. We distinguish desired transfer of persistent identity evidence from undesired transfer of capture-specific state. This IConFace diagnostic evaluates the latter for pose, expression, and semantic layout by comparing each output with its paired GT and the supplied references. Let $d_{\mathrm{pose}}$ be the mean absolute error over the three InsightFace pose angles, $d_{\mathrm{expr}}$ the mean absolute difference over the normalized mouth-, eye-, and smile-related proxies, and $s_{\mathrm{par}}$ the grouped ParseNet mIoU. Let $\bar d_i^{(k)}$ and $\bar s_i^{(\mathrm{par})}$ denote the corresponding output-to-reference measurements averaged over $R\in\mathcal{R}_i$. We define the target margins
\begin{align}
\Delta_i^{(k)}
&=\bar d_i^{(k)}-d_k(\hat I_i,G_i),
&& k\in\{\mathrm{pose},\mathrm{expr}\}, \\
\Delta_i^{(\mathrm{par})}
&=s_{\mathrm{par}}(\hat I_i,G_i)
-\bar s_i^{(\mathrm{par})}.
\end{align}
A positive margin means that the output is closer to the paired target than to the sample-balanced mean reference state. Averaging reference measurements within each sample before aggregation prevents samples with more references from receiving greater weight. We report the median margin and the percentages of target-preferred, reference-preferred, and tied samples.

Table~\ref{tab:supp_target_reference_alignment} yields positive median margins and majority target preference for every feature and dataset. On CelebA-Test-Ref and FFHQ-Ref Moderate, 91.2--100.0\% of samples are target-preferred. Under Severe degradation, target preference remains 95.0\% for pose, 94.9\% for parsing, and 73.0\% for expression. These results show that IConFace remains predominantly aligned with the paired target for the measured pose, expression, and semantic-layout attributes while using reference identity evidence.

Figures~\ref{fig:supp_ref_celeba_test_ref}--\ref{fig:supp_ref_ffhq_severe} visualize how cross-capture reference evidence complements target-aligned pose and expression across the three degradation settings. Ref$_1$ is displayed as visual context; aggregate identity evaluation uses the method-specific reference sets, while Tables~\ref{tab:supp_paired_target_fidelity} and~\ref{tab:supp_target_reference_alignment} provide paired-target and state-alignment measures.

\begin{table}[!t]
    \centering
    \caption{Target-versus-reference state alignment of IConFace. Reference measurements are averaged per sample before comparison with the paired GT, so every sample contributes once. Positive margins indicate target preference. Pose is in degrees, expression uses interocular-normalized landmark errors, and parsing uses mIoU differences. Target/Ref/Tie report the percentages of samples assigned to each outcome.}
    \label{tab:supp_target_reference_alignment}
    {\supptablefont\setlength{\tabcolsep}{1.2pt}\renewcommand{\arraystretch}{0.84}
    \begin{tabular}{@{}lccc@{}}
        \toprule
        Feature & Median margin $\uparrow$ & Target (\%) $\uparrow$ & Ref / Tie (\%) \\
        \midrule
        \multicolumn{4}{@{}l}{\textit{CelebA-Test-Ref} ($N=2{,}533$)} \\
        Pose & 5.993 & 99.1 & 0.9 / 0.0 \\
        Expression & 0.0252 & 91.2 & 8.8 / 0.0 \\
        Parsing & 0.315 & 100.0 & 0.0 / 0.0 \\
        \midrule
        \multicolumn{4}{@{}l}{\textit{FFHQ-Ref Moderate} ($N=857$)} \\
        Pose & 7.641 & 98.9 & 1.1 / 0.0 \\
        Expression & 0.0280 & 95.3 & 4.7 / 0.0 \\
        Parsing & 0.360 & 99.9 & 0.1 / 0.0 \\
        \midrule
        \multicolumn{4}{@{}l}{\textit{FFHQ-Ref Severe} ($N=857$)} \\
        Pose & 5.963 & 95.0 & 5.0 / 0.0 \\
        Expression & 0.0124 & 73.0 & 27.0 / 0.0 \\
        Parsing & 0.203 & 94.9 & 5.1 / 0.0 \\
        \bottomrule
    \end{tabular}}
\end{table}

\newcommand{\suppFigRefCeleba}{%
\begin{figure*}[p]
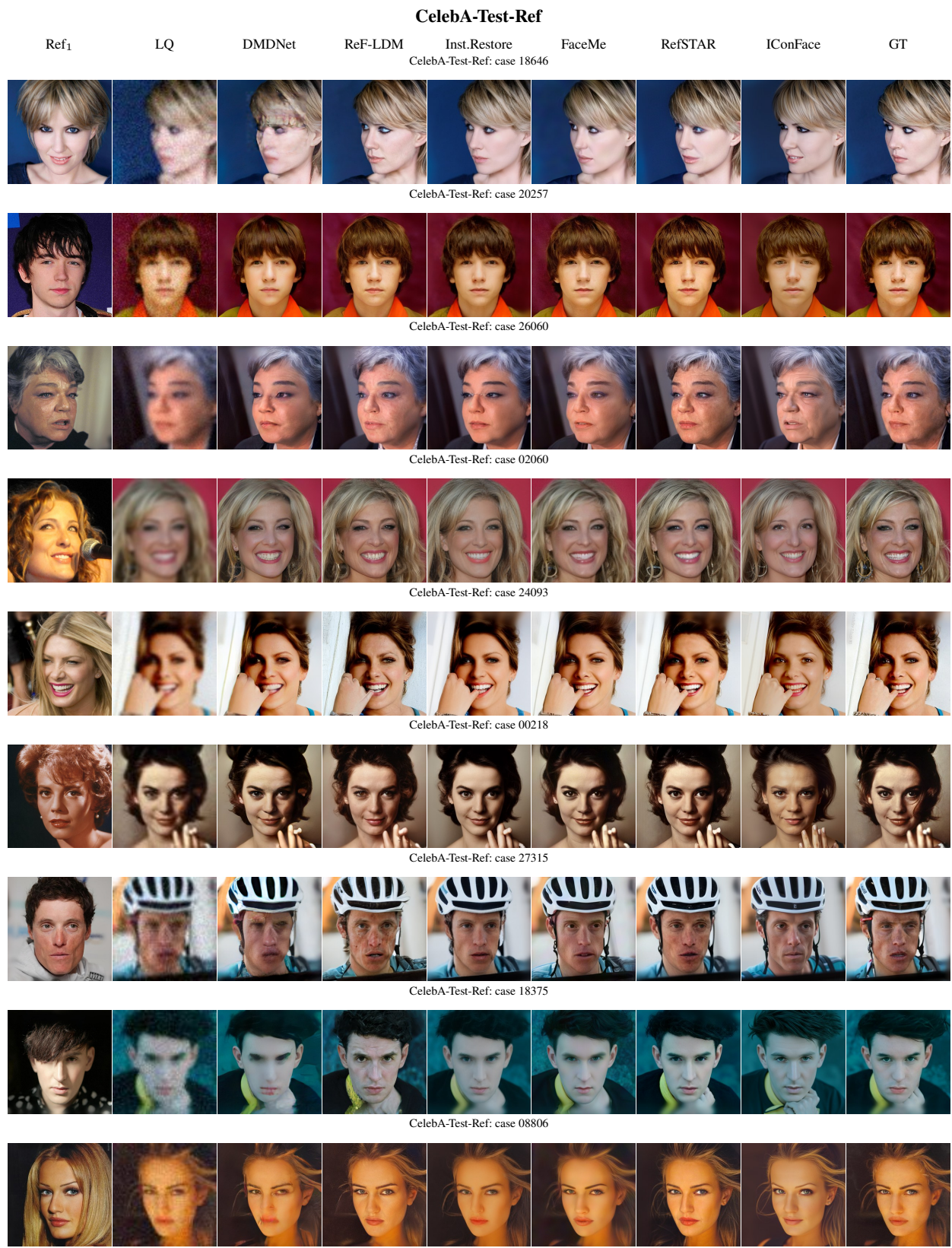

    \centering
    \datasetheader{CelebA-Test-Ref}
    \begingroup
    \qfigsetupcompact
    \begin{tabular}{@{}ccccccccc@{}}
    \qheadcompact{\supprefw}{Ref$_1$} &
    \qheadcompact{\supprefw}{LQ} &
    \qheadcompact{\supprefw}{DMDNet} &
    \qheadcompact{\supprefw}{ReF-LDM} &
    \qheadcompact{\supprefw}{Inst.Restore} &
    \qheadcompact{\supprefw}{FaceMe} &
    \qheadcompact{\supprefw}{RefSTAR} &
    \qheadcompact{\supprefw}{IConFace} &
    \qheadcompact{\supprefw}{GT}\\[-1pt]
    \supprefrow{CelebA-Test-Ref}{18646}{jpg}
    \supprefrow{CelebA-Test-Ref}{20257}{jpg}
    \supprefrow{CelebA-Test-Ref}{26060}{jpg}
    \supprefrow{CelebA-Test-Ref}{02060}{jpg}
    \supprefrow{CelebA-Test-Ref}{24093}{jpg}
    \supprefrow{CelebA-Test-Ref}{00218}{jpg}
    \supprefrow{CelebA-Test-Ref}{27315}{jpg}
    \supprefrow{CelebA-Test-Ref}{18375}{jpg}
    \supprefrow{CelebA-Test-Ref}{08806}{jpg}
    \end{tabular}
    \endgroup
    \caption{Reference-aware comparisons on nine CelebA-Test-Ref cases, including substantial Ref$_1$--GT cross-capture differences and broader restoration examples.}
    \label{fig:supp_ref_celeba_test_ref}
\end{figure*}
}

\newcommand{\suppFigRefModerate}{%
\begin{figure*}[p]
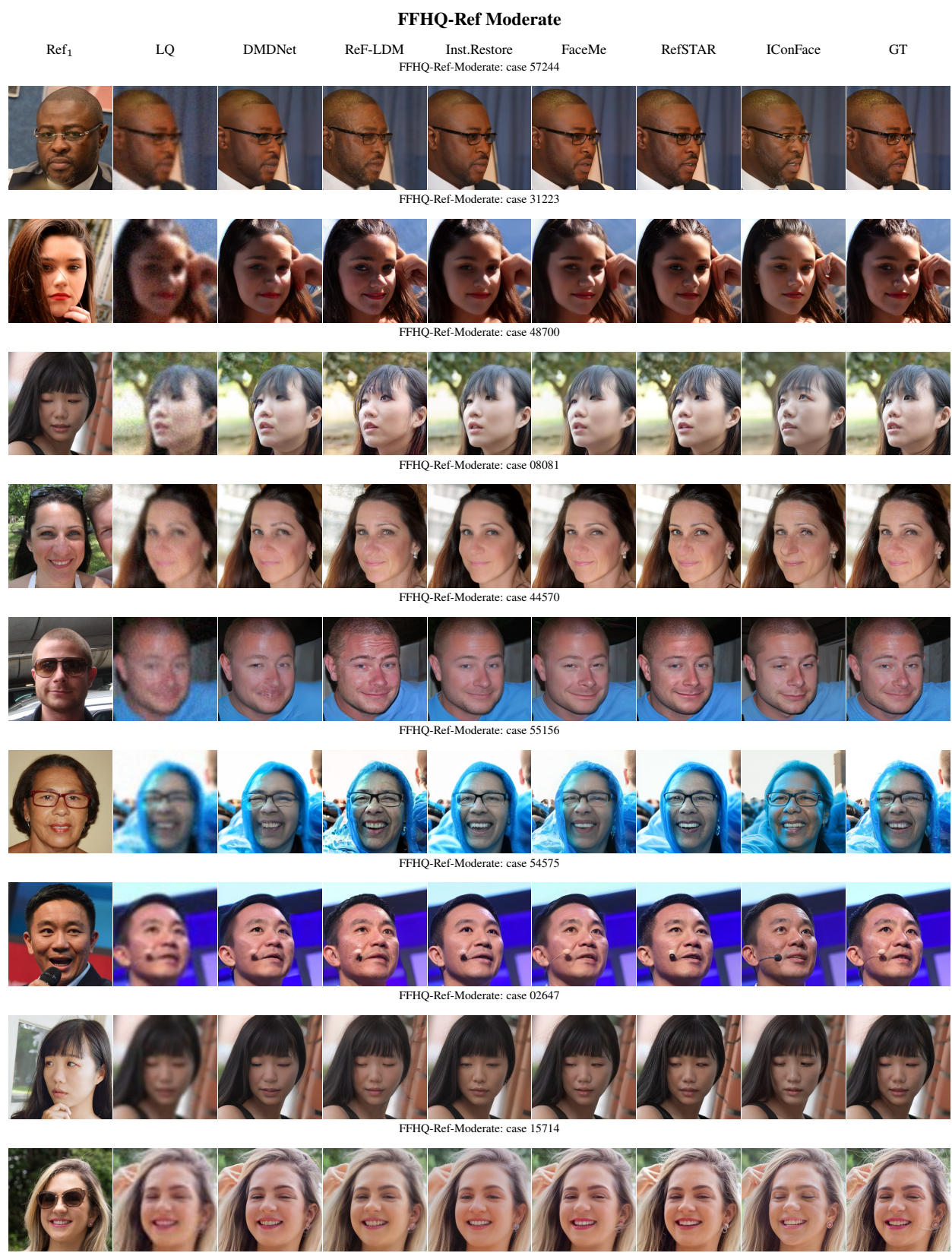

    \centering
    \datasetheader{FFHQ-Ref Moderate}
    \begingroup
    \qfigsetupcompact
    \begin{tabular}{@{}ccccccccc@{}}
    \qheadcompact{\supprefw}{Ref$_1$} &
    \qheadcompact{\supprefw}{LQ} &
    \qheadcompact{\supprefw}{DMDNet} &
    \qheadcompact{\supprefw}{ReF-LDM} &
    \qheadcompact{\supprefw}{Inst.Restore} &
    \qheadcompact{\supprefw}{FaceMe} &
    \qheadcompact{\supprefw}{RefSTAR} &
    \qheadcompact{\supprefw}{IConFace} &
    \qheadcompact{\supprefw}{GT}\\[-1pt]
    \supprefrow{FFHQ-Ref-Moderate}{57244}{jpg}
    \supprefrow{FFHQ-Ref-Moderate}{31223}{jpg}
    \supprefrow{FFHQ-Ref-Moderate}{48700}{jpg}
    \supprefrow{FFHQ-Ref-Moderate}{08081}{jpg}
    \supprefrow{FFHQ-Ref-Moderate}{44570}{jpg}
    \supprefrow{FFHQ-Ref-Moderate}{55156}{jpg}
    \supprefrow{FFHQ-Ref-Moderate}{54575}{jpg}
    \supprefrow{FFHQ-Ref-Moderate}{02647}{jpg}
    \supprefrow{FFHQ-Ref-Moderate}{15714}{jpg}
    \end{tabular}
    \endgroup
    \caption{Reference-aware comparisons on nine FFHQ-Ref Moderate cases combining Ref$_1$--GT cross-capture variation with representative restoration behavior.}
    \label{fig:supp_ref_ffhq_moderate}
\end{figure*}
}

\newcommand{\suppFigRefSevere}{%
\begin{figure*}[p]
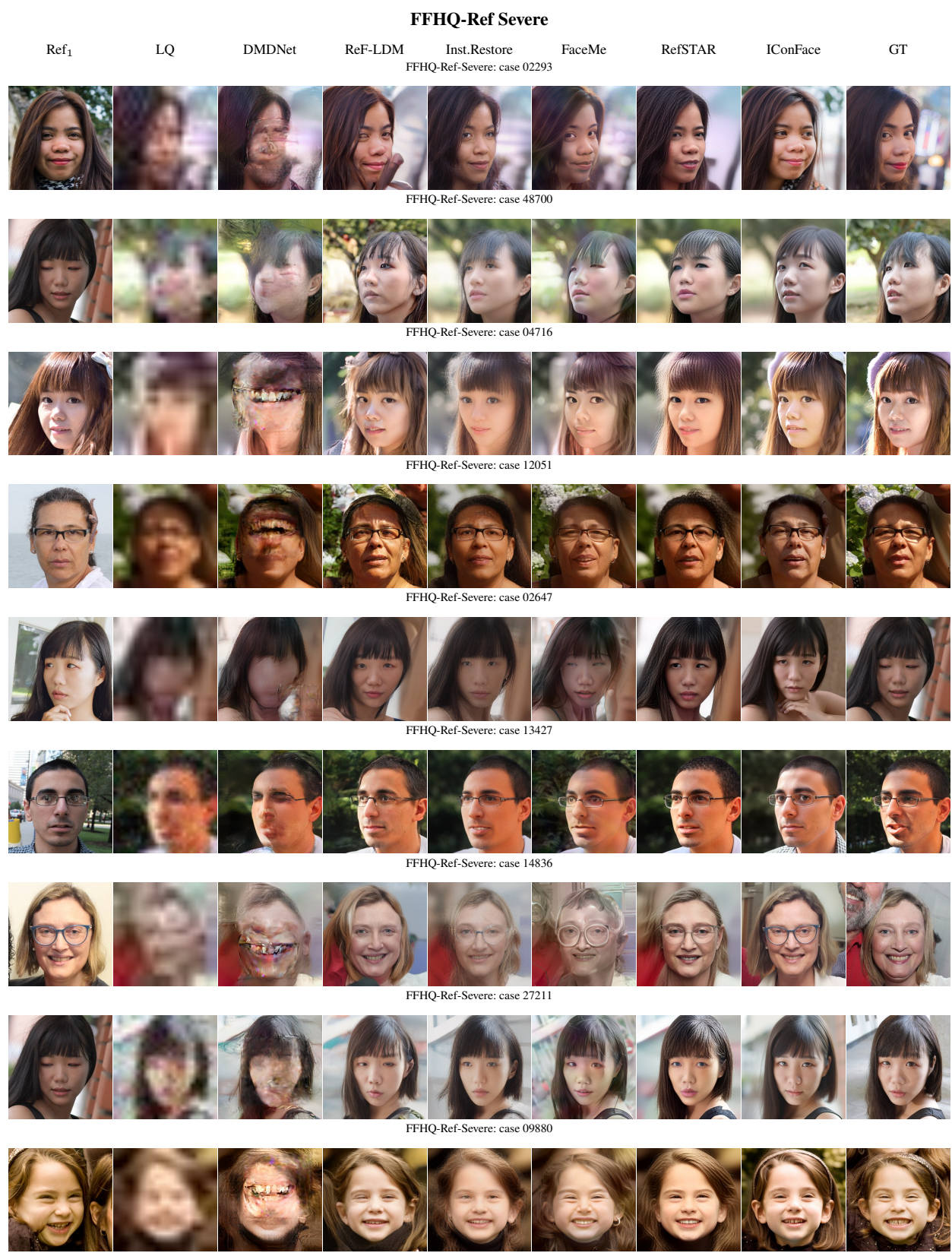

    \centering
    \datasetheader{FFHQ-Ref Severe}
    \begingroup
    \qfigsetupcompact
    \begin{tabular}{@{}ccccccccc@{}}
    \qheadcompact{\supprefw}{Ref$_1$} &
    \qheadcompact{\supprefw}{LQ} &
    \qheadcompact{\supprefw}{DMDNet} &
    \qheadcompact{\supprefw}{ReF-LDM} &
    \qheadcompact{\supprefw}{Inst.Restore} &
    \qheadcompact{\supprefw}{FaceMe} &
    \qheadcompact{\supprefw}{RefSTAR} &
    \qheadcompact{\supprefw}{IConFace} &
    \qheadcompact{\supprefw}{GT}\\[-1pt]
    \supprefrow{FFHQ-Ref-Severe}{02293}{jpg}
    \supprefrow{FFHQ-Ref-Severe}{48700}{jpg}
    \supprefrow{FFHQ-Ref-Severe}{04716}{jpg}
    \supprefrow{FFHQ-Ref-Severe}{12051}{jpg}
    \supprefrow{FFHQ-Ref-Severe}{02647}{jpg}
    \supprefrow{FFHQ-Ref-Severe}{13427}{jpg}
    \supprefrow{FFHQ-Ref-Severe}{14836}{jpg}
    \supprefrow{FFHQ-Ref-Severe}{27211}{jpg}
    \supprefrow{FFHQ-Ref-Severe}{09880}{jpg}
    \end{tabular}
    \endgroup
    \caption{Reference-aware comparisons on nine FFHQ-Ref Severe cases combining pronounced reference--target variation with severe degradations in which the observation retains less person-specific evidence.}
    \label{fig:supp_ref_ffhq_severe}
\end{figure*}
}

\FloatBarrier
\section{Localized Identity-Detail Evaluation}
\label{sec:supp_localized_detail}

\subsection{Benchmark Construction}
PID-167 evaluates persistent localized identity traits jointly supported by the paired GT and at least one same-identity reference. Qwen3-VL-4B-Instruct scans 2,533 CelebA-Test-Ref and 857 FFHQ-Ref Moderate GT images at 512 resolution using deterministic decoding, yielding 136 and 99 high-recall candidates. Eligible traits include localized moles, birthmarks, freckles, scars, stable pigmentation, dimples, cleft chins, and stable asymmetries; temporary or capture-dependent attributes such as hair, makeup, acne, and accessories are excluded. Cross-image human audit retains 82 and 85 cases, respectively, giving 167 cases whose primary trait is confirmed in the paired GT and at least one reference. The model-assisted stage proposes candidates only; retained traits are determined by human audit.

Each released case contains one primary audited trait with a localized description and recorded reference support. FFHQ-Ref Severe shares the same clean targets and is not duplicated as a second source split.

Each benchmark record specifies the source split and sample identifier, GT/reference mappings, trait location and audited description, checksums, and degraded-input path. Together, these fields uniquely identify the evaluation subset within the original benchmarks.

\subsection{Evaluation Protocol}
\begin{samepage}
For preliminary output assessment, Qwen3-VL-4B-Instruct receives only the paired GT, one restored output, and the audited trait description; neither references nor method names are included in the model input. Deterministic decoding is used. A trait is preserved only when the same localized feature remains visible in the corresponding facial region; unclear, erased, displaced, hallucinated, or heavily distorted traits are marked as not preserved. For primary feature $f_i$ and method $m$, the preservation rate is
\begin{equation}
    \mathrm{PID}(m)=\frac{1}{167}\sum_{i=1}^{167}\mathbf{1}\!\left[f_i\text{ is preserved by }m\right].
\end{equation}
\end{samepage}
Every reported binary judgment was subsequently reviewed manually.

\subsection{Results}
Table~\ref{tab:supp_mark_subset_construction} records the benchmark construction: human auditing retains 167 of 235 high-recall candidates, comprising 82 CelebA-Test-Ref cases and 85 FFHQ-Ref Moderate cases. The main paper reports the aggregate comparison over all 167 cases; Table~\ref{tab:supp_mark_diagnostic} resolves the result by source split.

\begin{table}[!t]
    \centering
    \caption{PID-167 construction and retained-set sources.}
    \label{tab:supp_mark_subset_construction}
    {\supptablefont\setlength{\tabcolsep}{3.0pt}\renewcommand{\arraystretch}{0.92}
    \begin{tabular}{@{}lr@{}}
        \toprule
        Item & Cases \\
        \midrule
        \multicolumn{2}{l}{\textit{Construction}} \\
        Qwen3-VL high-recall candidates & 235 \\
        Human-audited retained cases & 167 \\
        \midrule
        \multicolumn{2}{l}{\textit{Retained-set sources}} \\
        CelebA-Test-Ref & 82 \\
        FFHQ-Ref Moderate 512 & 85 \\
        \textbf{Total unique cases} & \textbf{167} \\
        \bottomrule
    \end{tabular}}
\end{table}

\begin{table}[!t]
    \centering
    \caption{Human-verified PID-167 preservation by source split. The aggregate 167-case comparison appears in main-paper Fig.~4; higher is better.}
    \label{tab:supp_mark_diagnostic}
    {\supptablefont\setlength{\tabcolsep}{1.4pt}\renewcommand{\arraystretch}{0.86}
    \begin{tabular}{@{}llrrr@{}}
        \toprule
        Split & Method & Cases & Preserved & \shortstack{Preservation\\Rate $\uparrow$} \\
        \midrule
        \multirow{6}{*}{CelebA-Test-Ref}
        & \mDMDNet & 82 & 63 & 0.768 \\
        & \mReFLDM & 82 & 74 & 0.902 \\
        & \mInstantRestore & 82 & 68 & 0.829 \\
        & \mFaceMe & 82 & 69 & 0.841 \\
        & \mRefSTAR & 82 & 75 & 0.915 \\
        & IConFace & 82 & 78 & \textbf{0.951} \\
        \midrule
        \multirow{6}{*}{FFHQ-Ref Moderate}
        & \mDMDNet & 85 & 63 & 0.741 \\
        & \mReFLDM & 85 & 71 & 0.835 \\
        & \mInstantRestore & 85 & 68 & 0.800 \\
        & \mFaceMe & 85 & 70 & 0.824 \\
        & \mRefSTAR & 85 & 79 & 0.929 \\
        & IConFace & 85 & 82 & \textbf{0.965} \\
        \bottomrule
    \end{tabular}}
\end{table}

IConFace preserves 78 of 82 traits on CelebA-Test-Ref (95.1\%) and 82 of 85 on FFHQ-Ref Moderate (96.5\%), yielding the highest observed rate on both source splits. Figure~\ref{fig:supp_visible_mark_examples} visualizes representative moles, scars, and pigmentation cues through enlarged red-box crops. In these selected cases, IConFace retains the reference-supported local cues more consistently than the displayed baselines, visually supporting the split-level quantitative comparison.

\newcommand{\suppFigPIDDetail}{%
\begin{figure*}[p]
    \centering
    \makebox[\textwidth][c]{\includegraphics[width=1.08\textwidth]{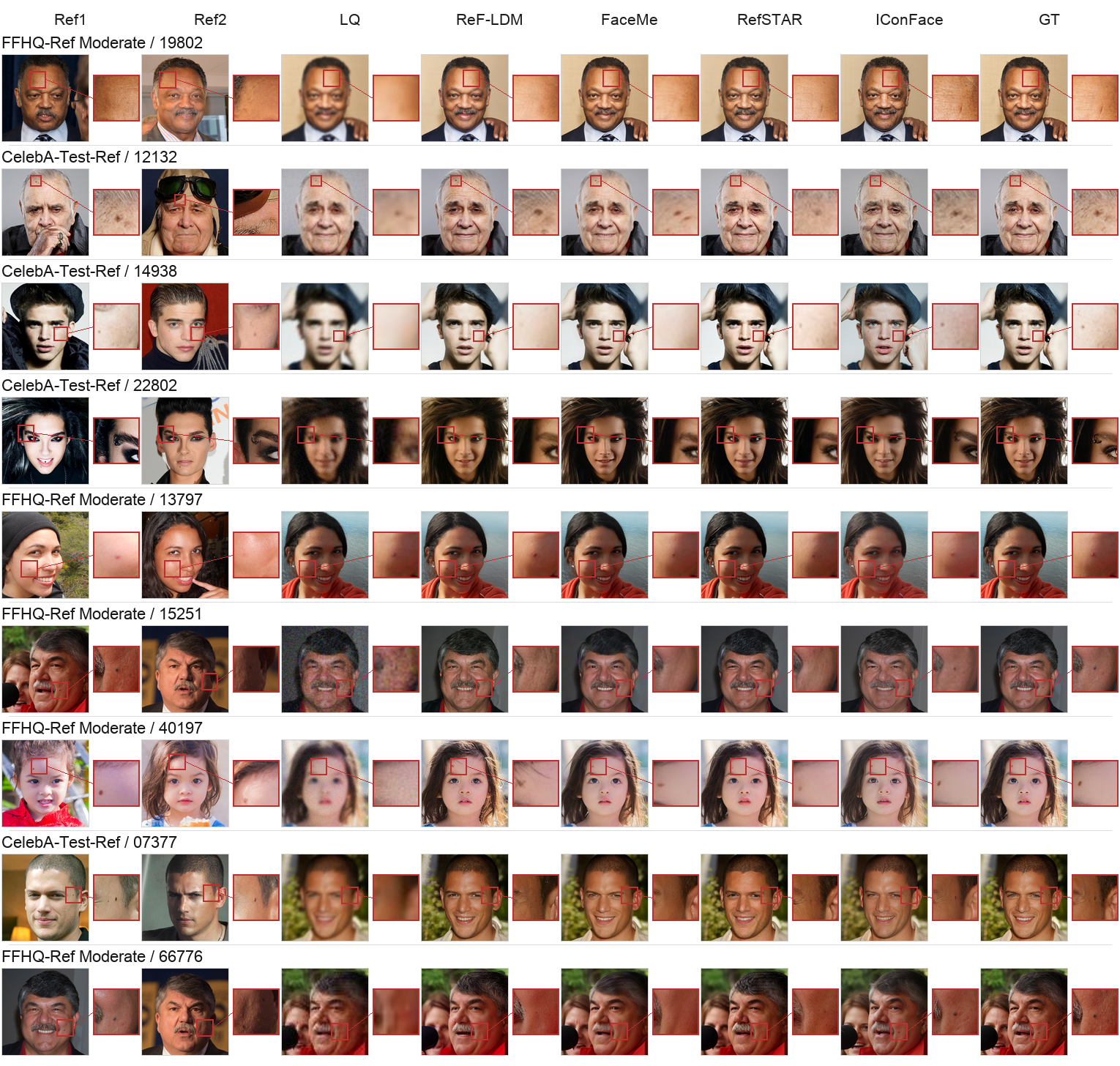}}
    \caption{Expanded localized identity-detail examples from PID-167. Columns show Ref$_1$, Ref$_2$, LQ, representative reference-aware outputs, IConFace, and GT. Each full image is paired with a local enlargement, with the red box marking its source region. References are shown here for audit context but are not included in the Qwen3-VL output-assessment input. Table~\ref{tab:supp_mark_diagnostic} reports the source-split quantitative comparison; the main paper reports aggregate rates.}
    \label{fig:supp_visible_mark_examples}
\end{figure*}
}

\FloatBarrier
\section{No-Reference Evaluation}
\label{sec:supp_no_reference}
Without references, the reference-token segment is omitted and the degraded image supplies the identity anchor. The main paper reports learned perceptual quality across five benchmarks; Table~\ref{tab:supp_celeba_test_distortion} complements that analysis with paired-target reconstruction on CelebA-Test.

\begin{table}[!t]
    \centering
    \caption{No-reference paired-target reconstruction on CelebA-Test.}
    \label{tab:supp_celeba_test_distortion}
    {\supptablefont\setlength{\tabcolsep}{3pt}\renewcommand{\arraystretch}{0.90}
    \begin{tabular}{lccc}
        \toprule
        Method & PSNR & SSIM & LPIPS \\
        \midrule
        \mCodeFormer & 25.146 & 0.685 & 0.227 \\
        \mGFPGAN & 25.105 & 0.696 & 0.241 \\
        \mVQFR & 23.233 & 0.658 & 0.244 \\
        \mRFpp & 25.313 & 0.685 & 0.226 \\
        \mDAEFR & 22.591 & 0.628 & 0.250 \\
        IConFace & 22.291 & 0.635 & 0.288 \\
        \bottomrule
    \end{tabular}}
\end{table}

Figures~\ref{fig:supp_noref_celeba_test}--\ref{fig:supp_noref_wider} provide no-reference comparisons on all five benchmarks, covering varied ages, poses, expressions, occlusions, illumination, compression, blur, and low resolution. Together, the galleries show consistent perceptual restoration across these domains.

\newcommand{\suppFigNoRefCeleba}{%
\begin{figure*}[p]
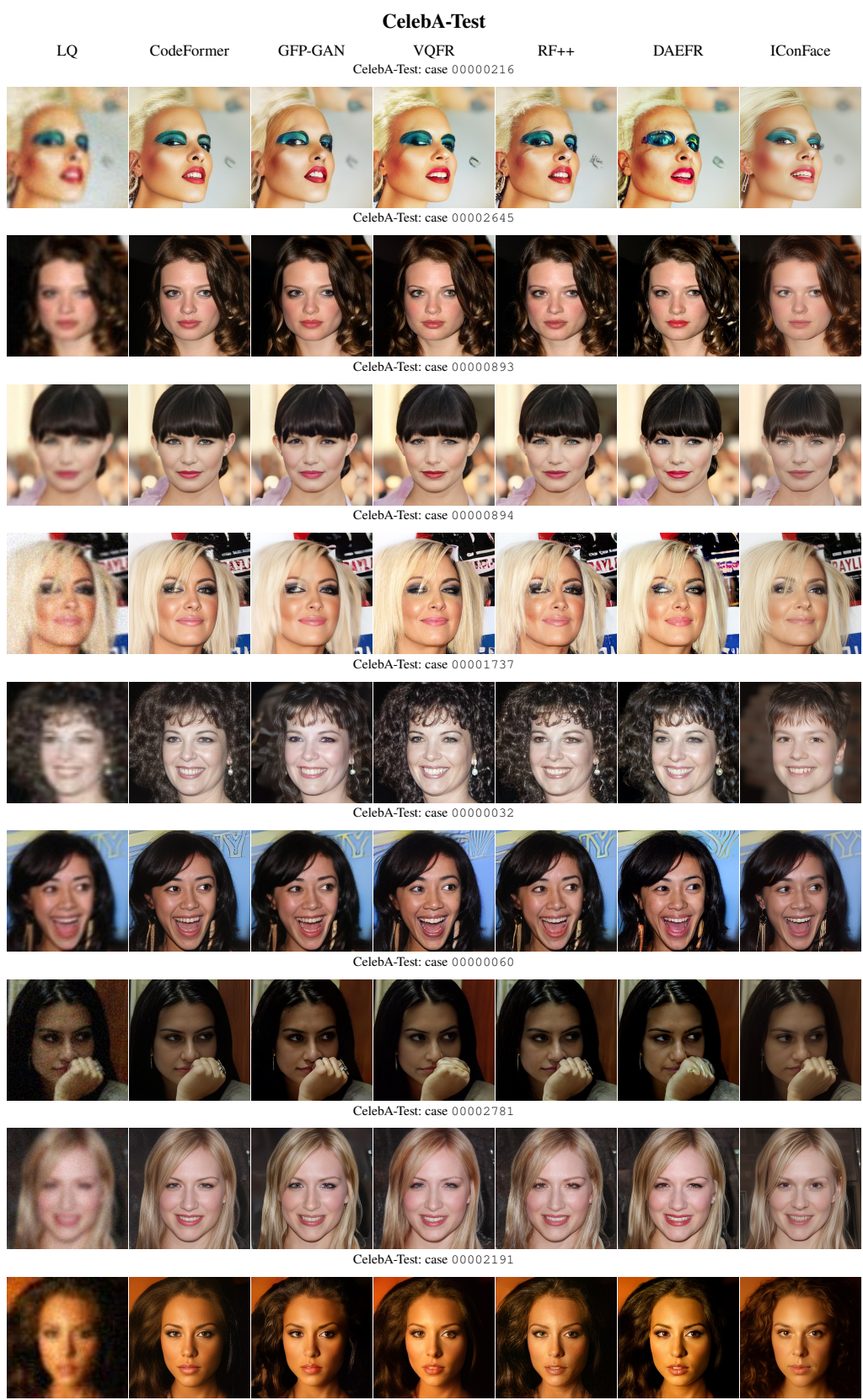

    \centering
    \datasetheader{CelebA-Test}
    \begingroup
    \qfigsetupblindcompact
    \begin{tabular}{@{}ccccccc@{}}
    \qheadcompact{\suppblindw}{LQ} &
    \qheadcompact{\suppblindw}{CodeFormer} &
    \qheadcompact{\suppblindw}{GFP-GAN} &
    \qheadcompact{\suppblindw}{VQFR} &
    \qheadcompact{\suppblindw}{RF++} &
    \qheadcompact{\suppblindw}{DAEFR} &
    \qheadcompact{\suppblindw}{IConFace}\\[-1pt]
    \suppblindrow{CelebA-Test}{\fingerroot/CelebA-Test}{00000216}
    \suppblindrow{CelebA-Test}{\fingerroot/CelebA-Test}{00002645}
    \suppblindrow{CelebA-Test}{\fingerroot/CelebA-Test}{00000893}
    \suppblindrow{CelebA-Test}{\fingerroot/CelebA-Test}{00000894}
    \suppblindrow{CelebA-Test}{fingers/candidates_100/no_reference/CelebA-Test}{00001737}
    \suppblindrow{CelebA-Test}{\fingerroot/CelebA-Test}{00000032}
    \suppblindrow{CelebA-Test}{\fingerroot/CelebA-Test}{00000060}
    \suppblindrow{CelebA-Test}{\fingerroot/CelebA-Test}{00002781}
    \suppblindrow{CelebA-Test}{fingers/candidates_100/no_reference/CelebA-Test}{00002191}
    \end{tabular}
    \endgroup
    \caption{Extended no-reference comparisons on nine paired CelebA-Test cases. Paired-target reconstruction metrics are reported in Table~\ref{tab:supp_celeba_test_distortion}.}
    \label{fig:supp_noref_celeba_test}
\end{figure*}
}

\newcommand{\suppFigNoRefLFW}{%
\begin{figure*}[p]
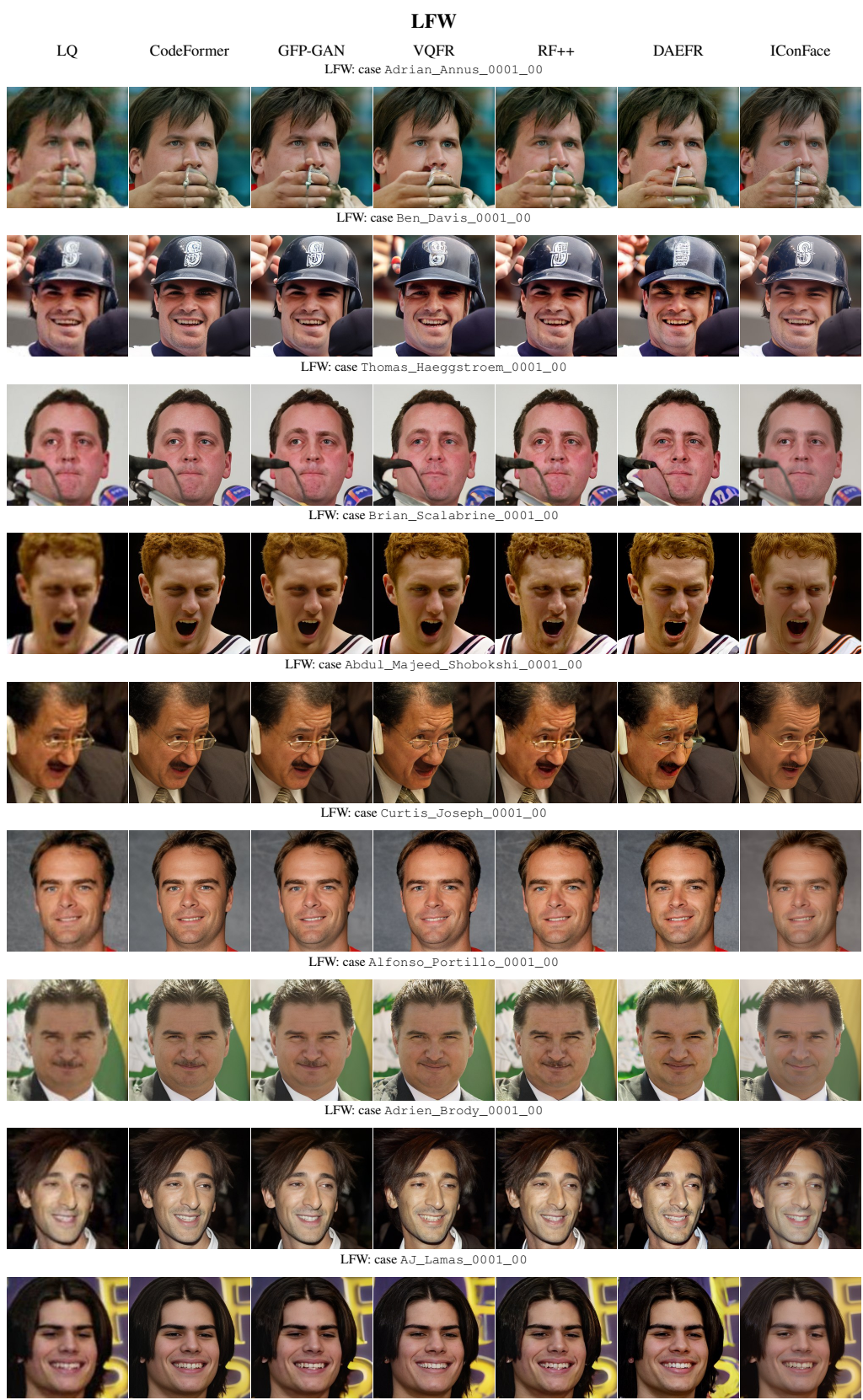

    \centering
    \datasetheadercompact{LFW}
    \begingroup
    \qfigsetupblindcompact
    \begin{tabular}{@{}ccccccc@{}}
    \qheadcompact{\suppblindw}{LQ} &
    \qheadcompact{\suppblindw}{CodeFormer} &
    \qheadcompact{\suppblindw}{GFP-GAN} &
    \qheadcompact{\suppblindw}{VQFR} &
    \qheadcompact{\suppblindw}{RF++} &
    \qheadcompact{\suppblindw}{DAEFR} &
    \qheadcompact{\suppblindw}{IConFace}\\[-1pt]
    \suppblindrow{LFW}{\fingerroot/LFW}{Adrian_Annus_0001_00}
    \suppblindrow{LFW}{\fingerroot/LFW}{Ben_Davis_0001_00}
    \suppblindrow{LFW}{\fingerroot/LFW}{Thomas_Haeggstroem_0001_00}
    \suppblindrow{LFW}{\fingerroot/LFW}{Brian_Scalabrine_0001_00}
    \suppblindrow{LFW}{\fingerroot/LFW}{Abdul_Majeed_Shobokshi_0001_00}
    \suppblindrow{LFW}{\fingerroot/LFW}{Curtis_Joseph_0001_00}
    \suppblindrow{LFW}{\fingerroot/LFW}{Alfonso_Portillo_0001_00}
    \suppblindrow{LFW}{\fingerroot/LFW}{Adrien_Brody_0001_00}
    \suppblindrow{LFW}{fingers/candidates_100/no_reference/LFW}{AJ_Lamas_0001_00}
    \end{tabular}
    \endgroup
    \caption{Extended no-reference comparisons on nine LFW cases spanning varied pose, expression, occlusion, and image quality.}
    \label{fig:supp_noref_lfw}
\end{figure*}
}

\newcommand{\suppFigNoRefCelebChild}{%
\begin{figure*}[p]
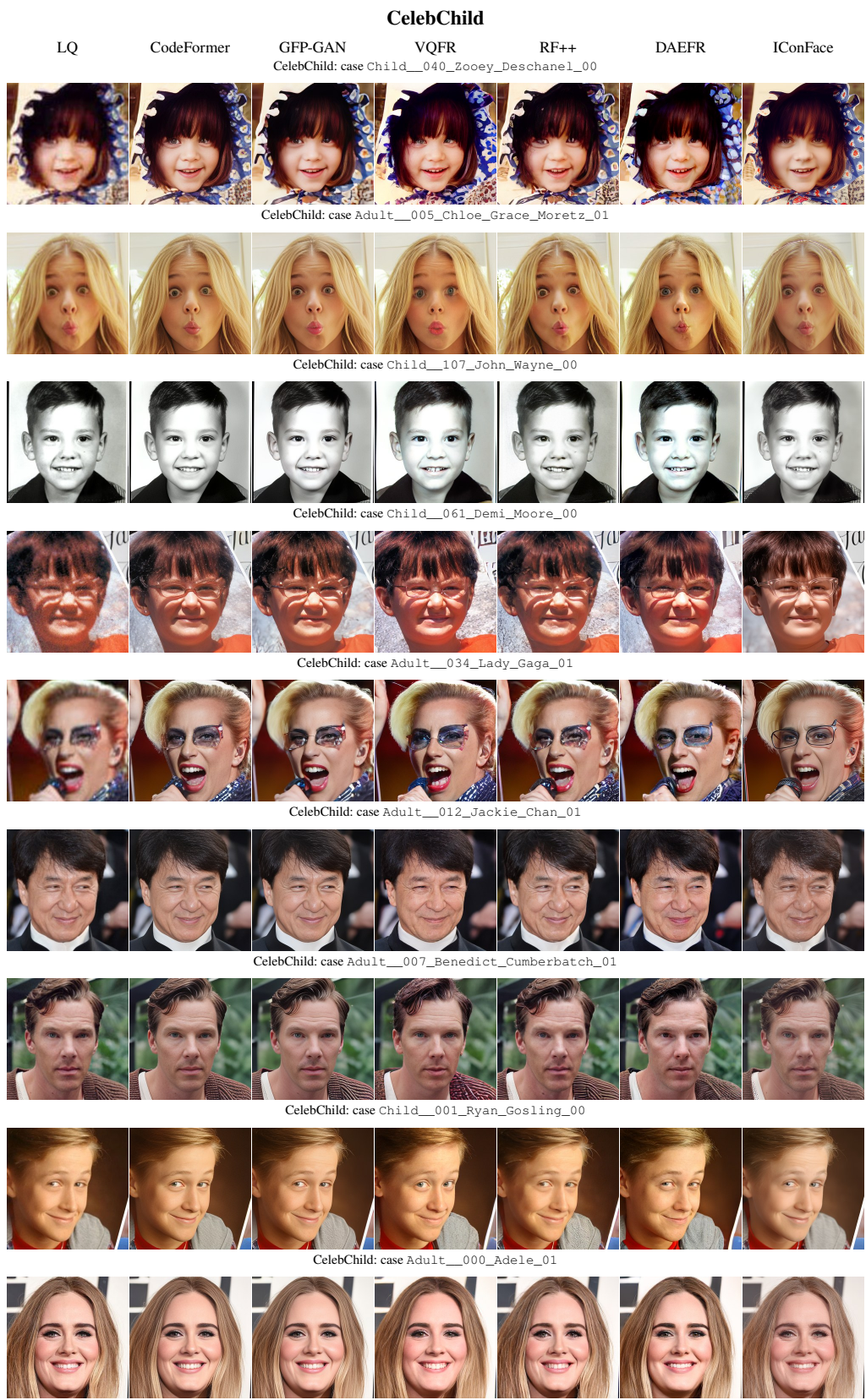

    \centering
    \datasetheader{CelebChild}
    \begingroup
    \qfigsetupblindcompact
    \begin{tabular}{@{}ccccccc@{}}
    \qheadcompact{\suppblindw}{LQ} &
    \qheadcompact{\suppblindw}{CodeFormer} &
    \qheadcompact{\suppblindw}{GFP-GAN} &
    \qheadcompact{\suppblindw}{VQFR} &
    \qheadcompact{\suppblindw}{RF++} &
    \qheadcompact{\suppblindw}{DAEFR} &
    \qheadcompact{\suppblindw}{IConFace}\\[-1pt]
    \suppblindrow{CelebChild}{\fingerroot/CelebChild}{Child__040_Zooey_Deschanel_00}
    \suppblindrow{CelebChild}{\fingerroot/CelebChild}{Adult__005_Chloe_Grace_Moretz_01}
    \suppblindrow{CelebChild}{\fingerroot/CelebChild}{Child__107_John_Wayne_00}
    \suppblindrow{CelebChild}{\fingerroot/CelebChild}{Child__061_Demi_Moore_00}
    \suppblindrow{CelebChild}{\fingerroot/CelebChild}{Adult__034_Lady_Gaga_01}
    \suppblindrow{CelebChild}{\fingerroot/CelebChild}{Adult__012_Jackie_Chan_01}
    \suppblindrow{CelebChild}{\fingerroot/CelebChild}{Adult__007_Benedict_Cumberbatch_01}
    \suppblindrow{CelebChild}{\fingerroot/CelebChild}{Child__001_Ryan_Gosling_00}
    \suppblindrow{CelebChild}{fingers/candidates_100/no_reference/CelebChild}{Adult__000_Adele_01}
    \end{tabular}
    \endgroup
    \caption{Extended no-reference comparisons on nine CelebChild cases covering both child and adult images.}
    \label{fig:supp_noref_celebchild}
\end{figure*}
}

\newcommand{\suppFigNoRefWebPhoto}{%
\begin{figure*}[p]
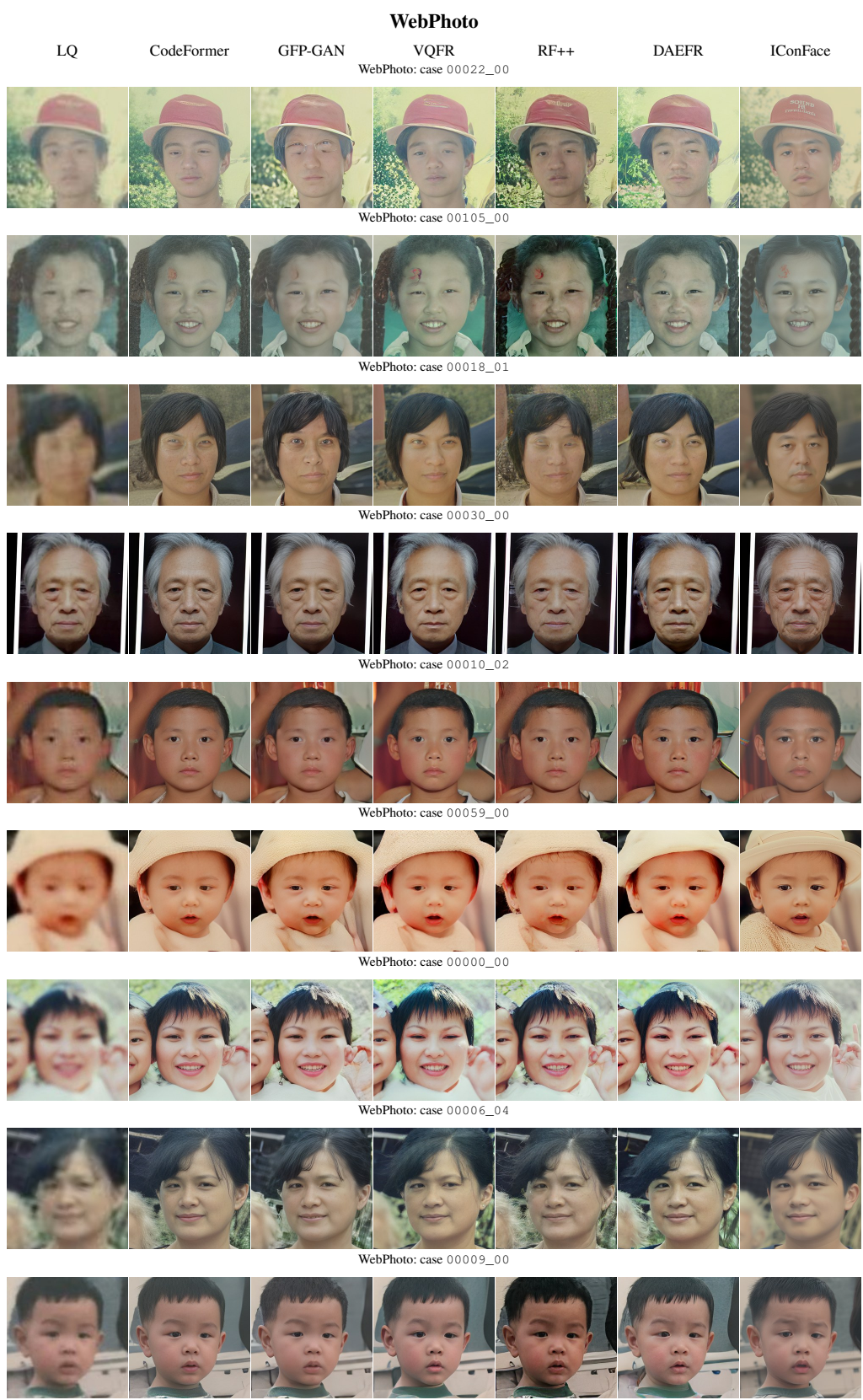

    \centering
    \datasetheadercompact{WebPhoto}
    \begingroup
    \qfigsetupblindcompact
    \begin{tabular}{@{}ccccccc@{}}
    \qheadcompact{\suppblindw}{LQ} &
    \qheadcompact{\suppblindw}{CodeFormer} &
    \qheadcompact{\suppblindw}{GFP-GAN} &
    \qheadcompact{\suppblindw}{VQFR} &
    \qheadcompact{\suppblindw}{RF++} &
    \qheadcompact{\suppblindw}{DAEFR} &
    \qheadcompact{\suppblindw}{IConFace}\\[-1pt]
    \suppblindrow{WebPhoto}{\fingerroot/WebPhoto}{00022_00}
    \suppblindrow{WebPhoto}{\fingerroot/WebPhoto}{00105_00}
    \suppblindrow{WebPhoto}{\fingerroot/WebPhoto}{00018_01}
    \suppblindrow{WebPhoto}{\fingerroot/WebPhoto}{00030_00}
    \suppblindrow{WebPhoto}{\fingerroot/WebPhoto}{00010_02}
    \suppblindrow{WebPhoto}{\fingerroot/WebPhoto}{00059_00}
    \suppblindrow{WebPhoto}{\fingerroot/WebPhoto}{00000_00}
    \suppblindrow{WebPhoto}{\fingerroot/WebPhoto}{00006_04}
    \suppblindrow{WebPhoto}{\fingerroot/WebPhoto}{00009_00}
    \end{tabular}
    \endgroup
    \caption{Extended no-reference comparisons on nine WebPhoto cases with low resolution, illumination variation, color shifts, and compression artifacts.}
    \label{fig:supp_noref_webphoto}
\end{figure*}
}

\newcommand{\suppFigNoRefWider}{%
\begin{figure*}[p]
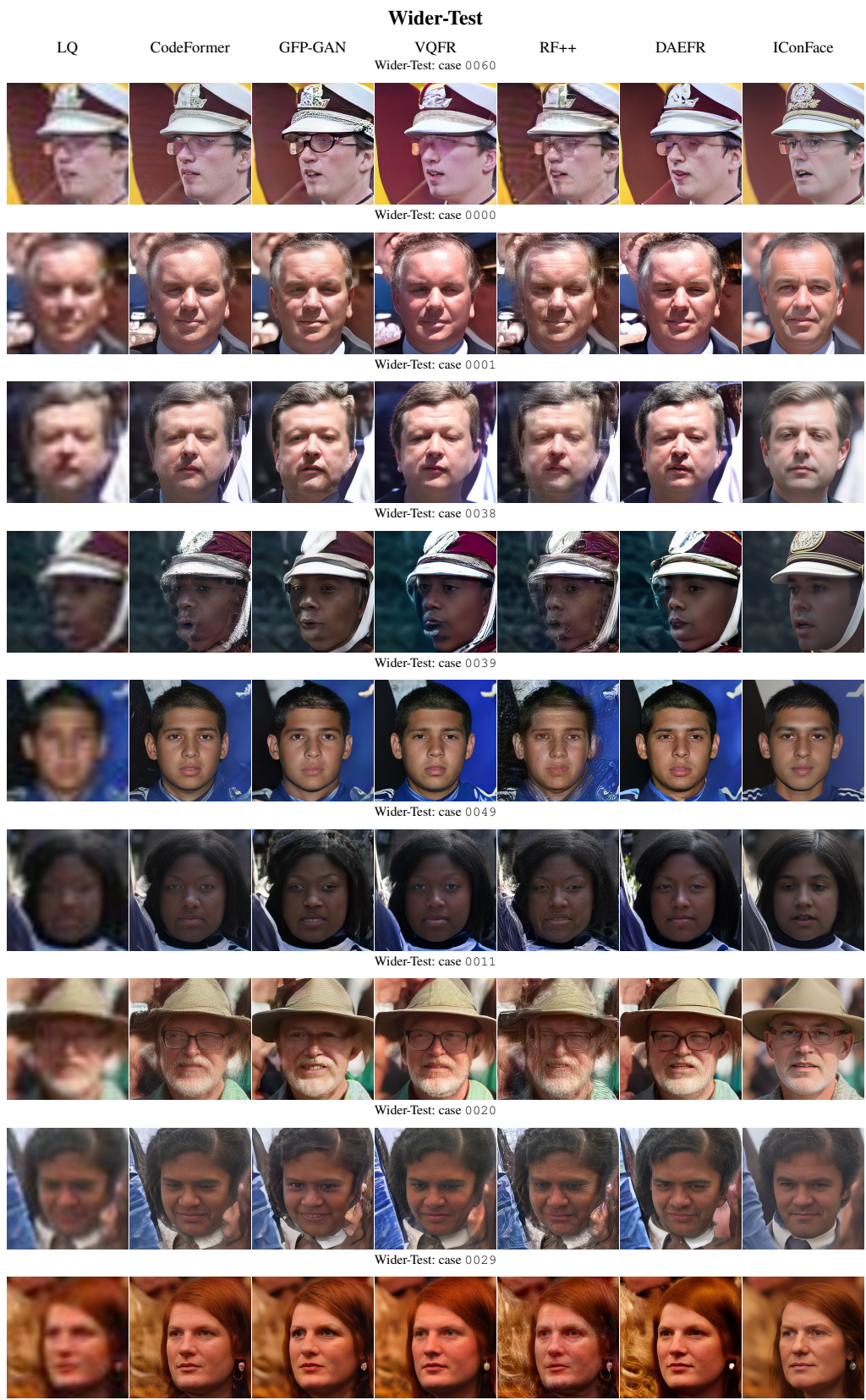

    \centering
    \datasetheader{Wider-Test}
    \begingroup
    \qfigsetupblindcompact
    \begin{tabular}{@{}ccccccc@{}}
    \qheadcompact{\suppblindw}{LQ} &
    \qheadcompact{\suppblindw}{CodeFormer} &
    \qheadcompact{\suppblindw}{GFP-GAN} &
    \qheadcompact{\suppblindw}{VQFR} &
    \qheadcompact{\suppblindw}{RF++} &
    \qheadcompact{\suppblindw}{DAEFR} &
    \qheadcompact{\suppblindw}{IConFace}\\[-1pt]
    \suppblindrow{Wider-Test}{\fingerroot/Wider-Test}{0060}
    \suppblindrow{Wider-Test}{\fingerroot/Wider-Test}{0000}
    \suppblindrow{Wider-Test}{\fingerroot/Wider-Test}{0001}
    \suppblindrow{Wider-Test}{\fingerroot/Wider-Test}{0038}
    \suppblindrow{Wider-Test}{fingers/candidates_100/no_reference/Wider-Test}{0039}
    \suppblindrow{Wider-Test}{fingers/candidates_100/no_reference/Wider-Test}{0049}
    \suppblindrow{Wider-Test}{\fingerroot/Wider-Test}{0011}
    \suppblindrow{Wider-Test}{\fingerroot/Wider-Test}{0020}
    \suppblindrow{Wider-Test}{fingers/candidates_100/no_reference/Wider-Test}{0029}
    \end{tabular}
    \endgroup
    \caption{Extended no-reference comparisons on nine Wider-Test cases with strong blur, low resolution, and pose variation.}
    \label{fig:supp_noref_wider}
\end{figure*}
}

\FloatBarrier
\section{Ablation Analysis}
\label{sec:supp_ablation}
Figures~\ref{fig:supp_ablation_moderate} and~\ref{fig:supp_ablation_severe} extend main-paper Table~4 and Fig.~7 with full-page comparisons of the five component configurations on FFHQ-Ref Moderate and Severe. Lower-right Arc labels report ArcFace similarity to the displayed Ref$_1$.

\newcommand{\suppFigAblationModerate}{%
\begin{figure*}[p]
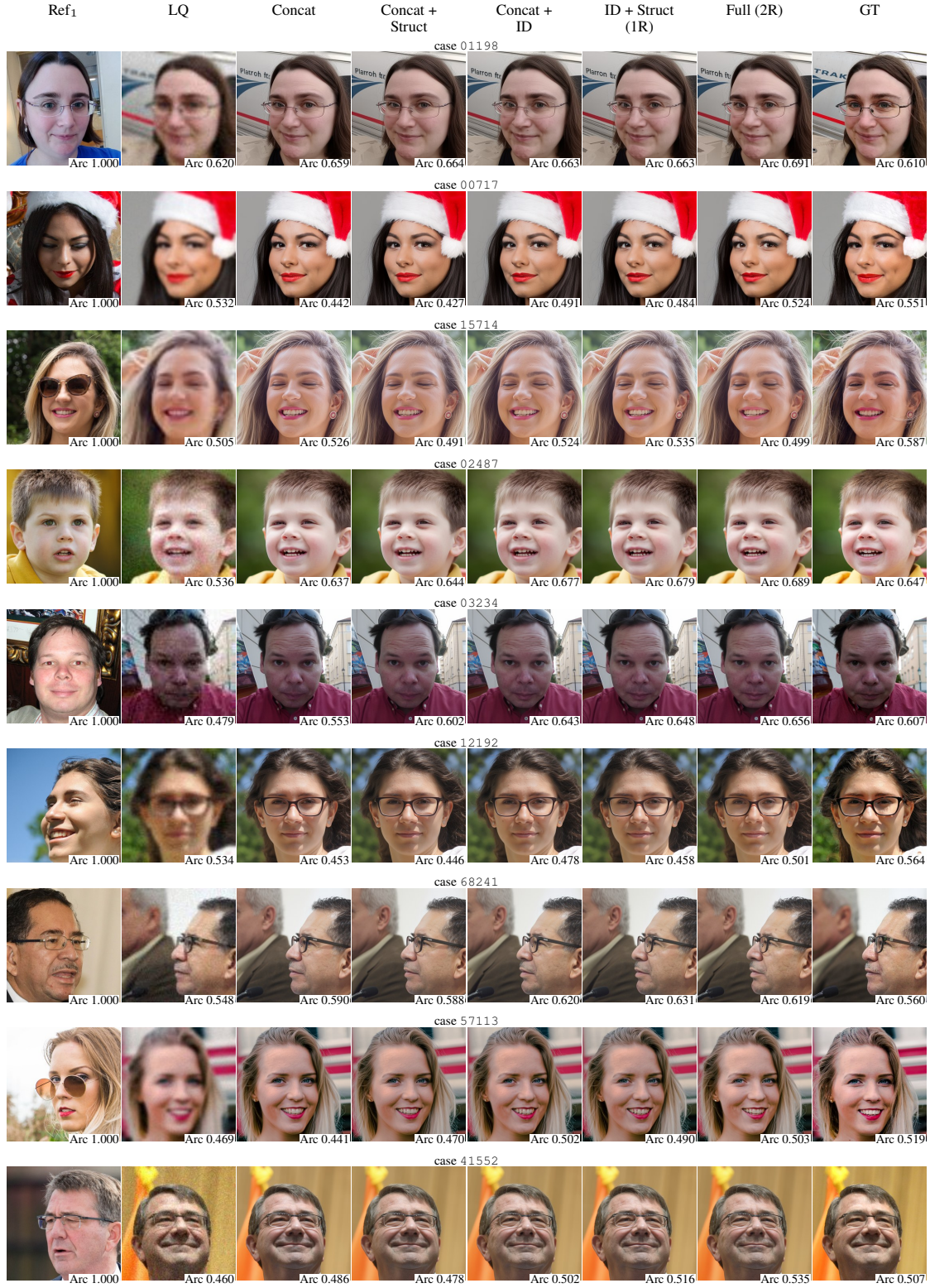

    \centering
    \datasetheader{FFHQ-Ref Moderate component ablations}
    \begingroup
    \qfigsetupcompact
    \begin{tabular}{@{}cccccccc@{}}
    \qheadcompact{\suppablw}{Ref$_1$} &
    \qheadcompact{\suppablw}{LQ} &
    \qheadcompact{\suppablw}{Concat} &
    \qheadcompact{\suppablw}{\shortstack{Concat +\\Struct}} &
    \qheadcompact{\suppablw}{\shortstack{Concat +\\ID}} &
    \qheadcompact{\suppablw}{\shortstack{ID + Struct\\(1R)}} &
    \qheadcompact{\suppablw}{Full (2R)} &
    \qheadcompact{\suppablw}{GT}\\[-1pt]
    \suppablscoredrow{fingers/candidates_100/ablation/FFHQ-Ref-Moderate}{01198}{0.620}{0.659}{0.664}{0.663}{0.663}{0.691}{0.610}
    \suppablscoredrow{fingers/iconface_ablation_top10/FFHQ-Ref-Moderate}{00717}{0.532}{0.442}{0.427}{0.491}{0.484}{0.524}{0.551}
    \suppablscoredrow{fingers/iconface_ablation_top10/FFHQ-Ref-Moderate}{15714}{0.505}{0.526}{0.491}{0.524}{0.535}{0.499}{0.587}
    \suppablscoredrow{fingers/candidates_100/ablation/FFHQ-Ref-Moderate}{02487}{0.536}{0.637}{0.644}{0.677}{0.679}{0.689}{0.647}
    \suppablscoredrow{fingers/candidates_100/ablation/FFHQ-Ref-Moderate}{03234}{0.479}{0.553}{0.602}{0.643}{0.648}{0.656}{0.607}
    \suppablscoredrow{fingers/iconface_ablation_top10/FFHQ-Ref-Moderate}{12192}{0.534}{0.453}{0.446}{0.478}{0.458}{0.501}{0.564}
    \suppablscoredrow{fingers/iconface_ablation_top10/FFHQ-Ref-Moderate}{68241}{0.548}{0.590}{0.588}{0.620}{0.631}{0.619}{0.560}
    \suppablscoredrow{fingers/iconface_ablation_top10/FFHQ-Ref-Moderate}{57113}{0.469}{0.441}{0.470}{0.502}{0.490}{0.503}{0.519}
    \suppablscoredrow{fingers/iconface_ablation_top10/FFHQ-Ref-Moderate}{41552}{0.460}{0.486}{0.478}{0.502}{0.516}{0.535}{0.507}
    \end{tabular}
    \endgroup
    \caption{Qualitative component ablations on nine FFHQ-Ref Moderate cases. The five output columns correspond to the configurations evaluated quantitatively in the main paper. Arc labels report ArcFace cosine similarity to the displayed Ref$_1$.}
    \label{fig:supp_ablation_moderate}
\end{figure*}
}

\newcommand{\suppFigAblationSevere}{%
\begin{figure*}[p]
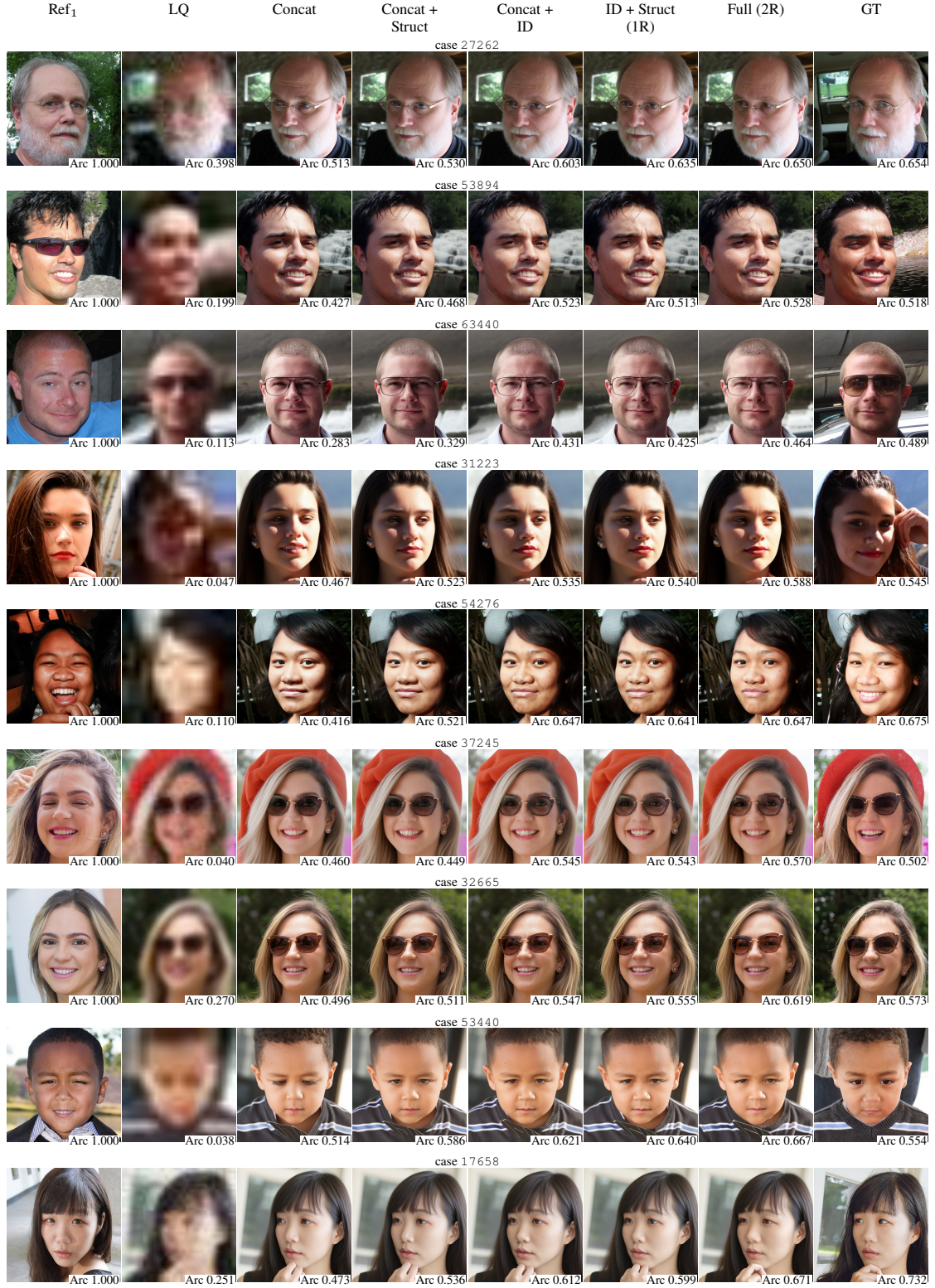

    \centering
    \datasetheader{FFHQ-Ref Severe component ablations}
    \begingroup
    \qfigsetupcompact
    \begin{tabular}{@{}cccccccc@{}}
    \qheadcompact{\suppablw}{Ref$_1$} &
    \qheadcompact{\suppablw}{LQ} &
    \qheadcompact{\suppablw}{Concat} &
    \qheadcompact{\suppablw}{\shortstack{Concat +\\Struct}} &
    \qheadcompact{\suppablw}{\shortstack{Concat +\\ID}} &
    \qheadcompact{\suppablw}{\shortstack{ID + Struct\\(1R)}} &
    \qheadcompact{\suppablw}{Full (2R)} &
    \qheadcompact{\suppablw}{GT}\\[-1pt]
    \suppablscoredrow{fingers/candidates_100/ablation/FFHQ-Ref-Severe}{27262}{0.398}{0.513}{0.530}{0.603}{0.635}{0.650}{0.654}
    \suppablscoredrow{fingers/candidates_100/ablation/FFHQ-Ref-Severe}{53894}{0.199}{0.427}{0.468}{0.523}{0.513}{0.528}{0.518}
    \suppablscoredrow{fingers/iconface_ablation_top10/FFHQ-Ref-Severe}{63440}{0.113}{0.283}{0.329}{0.431}{0.425}{0.464}{0.489}
    \suppablscoredrow{fingers/candidates_100/ablation/FFHQ-Ref-Severe}{31223}{0.047}{0.467}{0.523}{0.535}{0.540}{0.588}{0.545}
    \suppablscoredrow{fingers/candidates_100/ablation/FFHQ-Ref-Severe}{54276}{0.110}{0.416}{0.521}{0.647}{0.641}{0.647}{0.675}
    \suppablscoredrow{fingers/candidates_100/ablation/FFHQ-Ref-Severe}{37245}{0.040}{0.460}{0.449}{0.545}{0.543}{0.570}{0.502}
    \suppablscoredrow{fingers/candidates_100/ablation/FFHQ-Ref-Severe}{32665}{0.270}{0.496}{0.511}{0.547}{0.555}{0.619}{0.573}
    \suppablscoredrow{fingers/candidates_100/ablation/FFHQ-Ref-Severe}{53440}{0.038}{0.514}{0.586}{0.621}{0.640}{0.667}{0.554}
    \suppablscoredrow{fingers/iconface_ablation_top10/FFHQ-Ref-Severe}{17658}{0.251}{0.473}{0.536}{0.612}{0.599}{0.671}{0.732}
    \end{tabular}
    \endgroup
    \caption{Qualitative component ablations on nine FFHQ-Ref Severe cases. Separating the Severe examples from Fig.~\ref{fig:supp_ablation_moderate} makes local identity details and structure changes easier to inspect. Arc labels report ArcFace cosine similarity to the displayed Ref$_1$.}
    \label{fig:supp_ablation_severe}
\end{figure*}
}

\FloatBarrier
\suppFigRefCeleba
\FloatBarrier
\suppFigRefModerate
\FloatBarrier
\suppFigRefSevere
\FloatBarrier
\suppFigPIDDetail
\FloatBarrier
\suppFigNoRefCeleba
\FloatBarrier
\suppFigNoRefLFW
\FloatBarrier
\suppFigNoRefCelebChild
\FloatBarrier
\suppFigNoRefWebPhoto
\FloatBarrier
\suppFigNoRefWider
\FloatBarrier
\suppFigAblationModerate
\FloatBarrier
\suppFigAblationSevere
\FloatBarrier

\end{document}